\documentclass[lettersize,journal]{IEEEtran}

\usepackage[utf8]{inputenc}
\usepackage{epstopdf}
\usepackage[cmex10]{amsmath}
\usepackage{amsfonts}
\usepackage{amssymb}
\usepackage{mathrsfs}
\usepackage{graphicx}
\usepackage{verbatim}
\usepackage{array}
\usepackage{multirow}
\usepackage{multicol}
\usepackage{dcolumn}
\usepackage{color}
\usepackage[percent]{overpic}
\usepackage{mathtools}
\usepackage{arydshln}
\usepackage{url}
\usepackage{hyperref}

\usepackage{relsize}
\usepackage{booktabs}

\usepackage{algorithm}
\usepackage{algpseudocode}
\usepackage{caption}
\usepackage{subfigure}
\usepackage{subcaption}
\usepackage{lipsum}
\usepackage{bbm}
\usepackage{xcolor}

\DeclareGraphicsExtensions{.eps}
\AppendGraphicsExtensions{.pdf}

\DeclareMathOperator*{\argmax}{arg\,max}
\usepackage[noadjust]{cite}

\begin{document}
	
\title{Selecting Critical Scenarios of DER Adoption in Distribution Grids Using Bayesian Optimization}

\author{Olivier Mulkin,
	Miguel Heleno,~\IEEEmembership{Senior Member,~IEEE}, Mike Ludkovski
    
	\thanks{M. Heleno is with the Lawrence Berkeley National Laboratory, Berkeley, CA, U.S.A. (e-mail: MiguelHeleno@lbl.gov). O. Mulkin and M. Ludkovski are with the Department of Statistics \& Applied Probability, UC Santa Barbara}
}


\newcommand{\ml}[1]{\textcolor{red}{[ML: #1]}} 
\newcommand{\om}[1]{\textcolor{red}{[OM: #1]}} 
\newcommand{\edit}[1]{\textcolor{black}{#1}} 
\newcommand{\codecomment}[1]{\textcolor{gray}{#1}} 

\newcommand{\nodes}[0]{\mathcal{B}}
\newcommand{\lines}[0]{\mathcal{L}}
\newcommand{\scens}[0]{\mathcal{X}}
\newcommand{\adopters}[0]{\mathcal{A}}
\newcommand{\nadopt}[0]{A}

\newcommand{\bK}{\mathbf{K}}
\newcommand{\by}[0]{\mathbf{y}}
\newcommand{\bY}[0]{\mathbf{Y}}
\newcommand{\bx}[0]{\mathbf{x}}
\newcommand{\bX}[0]{\mathbf{X}}
\newcommand{\cX}[0]{\mathcal{X}}
\newcommand{\cO}[0]{\mathcal{O}}
\newcommand{\cD}[0]{\mathcal{D}}
\newcommand{\cS}[0]{\mathcal{S}}
\newcommand{\cP}[0]{\mathcal{P}}
\newcommand{\cC}[0]{\mathcal{C}}
\newcommand{\cE}[0]{\mathcal{E}}
\newcommand{\Nstop}[0]{n^\star}

\newcommand{\Xn}[1][n]{\ensuremath{\mathbf{X}_{#1}}}
\newcommand{\Xt}[0]{\ensuremath{\mathcal{X}_{t}}}
\newcommand{\datan}[0]{\mathcal{D}_{n}}
\newcommand{\Viol}[2][k]{\ensuremath{V_{#1}\left(#2\right)}}
\newcommand{\ind}[1]{\mathbbm{1}\left({#1}\right)}
\newcommand{\proba}[1]{\ensuremath{\mathbb{P}\!\left(#1\right)}}
\NewDocumentCommand{\expect}{m O{}}
{
    \ensuremath{\mathbb{E}_{#2}\!\left[#1 \right]}
}

\maketitle

\begin{abstract}
    We develop a new methodology to select scenarios of DER adoption most critical for distribution grids. Anticipating risks of future voltage and line flow violations due to additional PV adopters is central for utility investment planning but continues to rely on deterministic or ad hoc scenario selection. We propose a highly efficient search framework based on multi-objective Bayesian Optimization. We treat underlying grid stress metrics as computationally expensive black-box functions, approximated via Gaussian Process surrogates and design an acquisition function based on probability of scenarios being Pareto-critical across a collection of line- and bus-based violation objectives. Our approach provides a statistical guarantee and offers an order of magnitude speed-up relative to a conservative exhaustive search. Case studies on realistic feeders with 200-400 buses demonstrate the effectiveness and accuracy of our approach.

    
\end{abstract}

\begin{IEEEkeywords}
Distribution Planning, DER Adoption, Scenario Selection, Bayesian Optimization, Gaussian Process Surrogates
\end{IEEEkeywords}


\section{Introduction} \label{Introduction}

\IEEEPARstart{I}{nvestments} in new distribution grid capacity account for a significant share of utility capital expenditures \cite{EEI2019}, requiring utilities to periodically justify those investments to regulators through detailed assessments of distribution grid needs \cite{pge_gna_2023, sce_dsp_2021}. These assessments rely heavily on long-term forecasts (typically spanning 5 to 10 years) of distribution system demand and Distributed Energy Resource (DER) penetration \cite{epri_load_forecasting_2013}, which drive distribution capacity needs.

However, while there are methodologies to predict future DER diffusion at the city, substation, or neighborhood level \cite{Heymann2021}, understanding ``where" exactly these DERs will appear within a distribution feeder remains a challenge. In fact, for feeders with hundreds or even thousands of consumers, a single DER diffusion trajectory can translate into a high-dimensional set of scenarios of potential individual DER adoption at the nodes of the distribution grid. Planning localized feeder upgrades, such as reconductoring or voltage regulation, requires identifying, in this high-dimensional space, the specific combinations of nodal adoption patterns that drive new capacity needs.


To address this issue, this paper proposes a new methodology, based on Bayesian Optimization (BO), to forecast and identify scenarios of behind-the-meter DER adoption that are critical to distribution grid planning.  

\subsection{Literature Review}

Long-term forecasts of loads and DERs start at the system level to provide a broad picture of load and DER growth across the utility's territory while ensuring compatibility with other planning processes \cite{hawaiianelectric_dsp_2021, sce_dsp_2021}. Utilities capture data on historical consumption, demographic and economic indicators, weather trends, and DER policy adoption rates \cite{pge_gna_2023, pacificorp_dsp_2021} and use various statistical-based methodologies to generate such aggregate-level forecasts. For example, \cite{McSharry2005} introduces a probabilistic forecasting model to predict the magnitude and timing of peak electricity demand, \edit{\cite{EVANGELOPOULOS2022107847} applies spatial forecasting to assess the impact of load growth in power distribution networks}, \cite{Hyndman2010} proposes a density forecasting approach to address the uncertainty associated with long-term planning, and \cite{Hong2014} uses hourly data to model long-term trends in probabilistic forecasts. A methodology that uses an eXtreme Gradient Boosting algorithm with different sequential configurations is proposed in \cite{Zhang2024} to combine energy consumption and peak power demand forecasting. In a different direction, the work of \cite{Dong2020} proposes a hybrid approach that combines top-down features (economic and weather) with bottom-up features (individual consumer loads and DER adoption propensity) to provide long-term forecasts. \edit{More recently, \cite{Li2022} introduce a method based on system dynamics to forecast long-term electricity consumption by dividing the driving factors of energy use into population, per capita GDP, energy intensity, and energy structure. In \cite{Osaka2024} a hybrid convolutional neural network and long short-term memory method is proposed to forecast monthly energy consumption for three year planning horizons.}

For distribution system planning, utilities disaggregate these system-level forecasts into smaller geographic units using tools like LoadSEER \cite{hawaiianelectric_dsp_2021, pge_gna_2023}. These tools rely on statistical spatial regressions \cite{Bernards} or agent-based models \cite{ROBINSON2015273, sce_dsp_2021} to generate future scenarios of load growth and DER diffusion. In the literature, earlier works proposed land-use pattern recognition techniques to anticipate new demand using cellular data \cite{Hung-Chih2002, Carreno}, while recent studies focus on forecasting a combination of loads and DERs, including their temporal characteristics \cite{Ye2019, HEYMANN2019}. However, these models are often based on geographical representations (e.g., neighborhoods, zip codes), which do not necessarily correspond to specific feeder locations. \edit{To forecast household-level consumption, \cite{Botman2023} proposes an ensemble of forecasts based on the historical median distribution among similar households. The forecast of DERs at the consumer site} can be based on rules (e.g., assuming DER power proportional to base load \cite{RODRIGUEZCALVO2017121}) or on individual consumer models \cite{HELENO2020115826}. For example, \cite{Sun2021} develops a feeder-specific agent-based model integrating customer GIS data, EV adoption criteria (e.g., neighbor influence, car age), and charging station placement. This feeder-level adoption is highly granular, down to individual parcels and specific customer distribution transformers. In the utility space, Portland General Electric (PGE) developed its own AdopDER tool, a statistical model that captures consumer-level propensity to adopt DERs \cite{pgeoregon_dsp_2021}.

Although these DER forecasting methods provide feeder-level spatial granularity, they may not be suitable for planning purposes. This is because they primarily identify load and DER patterns that are more likely to occur, rather than those that would impose greater stress on the system. To address this limitation, recent works have proposed methods to select load growth and DER penetration scenarios that could trigger capacity upgrades by mapping these scenarios to specific equipment capacity violations. For instance, the authors of \cite{Li2020} presented a method to forecast substation annual maximum demand by analyzing extreme load events and their relationship with external factors, leveraging extreme value theory. Similarly, \cite{Dong2021} introduced a data-driven approach to estimate transformer ratings under varying temperature and load conditions. Regarding DER adoption, the authors of \cite{Sexauer} proposed a method based on a binomial distribution to predict the probability of transformer overloading caused by electric vehicle (EV) adoption. 

Nonetheless, a major limitation of these approaches is that they identify the most critical scenarios in relation to a single equipment (e.g., substations or service transformers). Expanding to the entire distribution system would require identifying all critical scenarios corresponding to all potential system-level violations. 
In distribution grids with \edit{hundreds} of nodes and millions of potential scenarios of DER adoption, this becomes a search in a high-dimensional space making exhaustive evaluation infeasible. \edit{We propose a guided ``smart'' search via Bayesian Optimization, designed to address this challenge by substantially reducing the number of scenarios that must be evaluated, as demonstrated in feeders with several hundred nodes. While we focus on these realistic test cases, the proposed approach offers a scalable foundation that could be extended to larger networks in future work.}

\subsection{Contributions}

This paper introduces a novel methodology for forecasting long-term DER adoption patterns for the purpose of distribution system planning. Unlike existing high-resolution locational DER forecast approaches \cite{RODRIGUEZCALVO2017121, HELENO2020115826, Sun2021, pgeoregon_dsp_2021}, which focus on identifying the most probable scenarios, our methodology prioritizes finding the scenarios that are most critical to the grid and, therefore, more relevant for planning exercises. Previous forecasting approaches following this logic have been limited to mapping critical scenarios based on the capacity of single distribution grid assets \cite{Li2020, Dong2021, Sexauer}. Our work extends this concept to a system-wide perspective, leveraging BO to navigate the high-dimensional space and identify critical scenarios relevant for planning upgrades in the entire grid.

Our main contributions are threefold:

\begin{itemize} 

\item We link long-term DER adoption scenarios to future violations of distribution grid limits (voltage and line capacity), in order to identify the set of critical scenarios that are most relevant for distribution grid planning. We map such critical scenarios to the Pareto frontier of the above grid stress objectives vector. 

\item We propose an algorithm that efficiently performs this search, provides statistical guarantees, and achieves an order-of-magnitude speed-up compared to a conservative exhaustive search. 

\item Using two realistic case studies, we demonstrate that the DER adoption scenarios critical for distribution grid planning are not necessarily those with extreme aggregated adoption patterns. This highlights the importance of conducting a thorough and systematic locational search, such as the one presented in this paper. 

\end{itemize}

Statistically, our methodology leverages Gaussian Process (GP) surrogates to efficiently forecast line and bus voltage violations. GPs have been previously used in several power system applications, for instance, to describe the uncertainty of optimal power flows \cite{pareek_gaussian_2021}, to perform transient stability analysis of large-scale power systems \cite{ye_physics-informed_2023, ye_high_2024}, to simulate renewable energy power generation  \cite{ludkovski_large_2022, li_probabilistic_2024}, etc. However, this paper is, to the best of our knowledge, the first application of GPs in long-term forecasts of distribution grid load and stress for the purpose of planning.

\subsection{Paper organization}
The rest of the paper is organized as follows. Sections II and III introduce the problem statement and the methodology, respectively. Numerical results on two realistic feeders are presented in Section IV; Section V discusses the performance of our model and Section VI concludes.


\section{Problem Statement} \label{sec:prob_statement}

A distribution planning process implies a grid needs assessment \cite{pge_gna_2023}, which is conducted in three main steps: i) define scenarios for load and DER penetration; ii) run a feeder-level power flow evaluation for those scenarios and assess potential line/transformer capacity and voltage violations; iii) propose a planning solution based on distribution capital projects to upgrade the system and solve those violations. The selection of scenarios in the first step is critical, as it directly influences the power flow analysis and ultimately shapes the planning solutions. In this section, we describe the problem of identifying the set of \textit{critical scenarios}, i.e.~those whose evaluation and resolution of violations ensures the grid remains feasible for all other potential scenarios. Throughout the paper, we use the behind-the-meter adoption of PV to better illustrate the problem and the methodology proposed. The extension to other types of DERs or to load is straightforward.

\subsection{Scenarios of Adoption}
We consider a radial distribution grid with $L$ lines and $B$ buses. Let $\adopters$ be the set of $\nadopt = |\adopters|$ potential PV adopters, i.e.~network nodes where a PV system can be installed. We simulate scenarios of PV adoption with an agent-based extension of the dGen diffusion model \cite{osti_1239054}, \edit{introduced in \cite{cordova_netload_2024}}. The probability of an agent adopting at time $t + \Delta t$ given the network state at time $t$ is
\begin{equation}\label{eq:bass_prob}
    \proba{a_{t + \Delta t, j} = 1 | a_{t, j} = 0} = 
    \left( p + \dfrac{q}{\nadopt} \sum_{k \in \adopters}a_{t, k} \right),
\end{equation}
where $p$ and $q$ are the coefficients of innovation and adoption respectively, and $a_{t, j} = 1$ if agent $j$ has adopted PV at time $t$, and $a_{t, j}=0$ otherwise. The initial adoption status $a_{0, j}$ 
is uniformly sampled with a probability matching currently observed adoption rates. For a given time horizon $T$, scenarios describing potential future PV adoption over the network \edit{are obtained by independent} Monte Carlo simulations with transition probabilities \eqref{eq:bass_prob}. 

We denote by $\cX$ the space of all scenarios based on model (\ref{eq:bass_prob}). Formally, $\cX=\{0,1\}^\nadopt$ as any scenario has a strictly positive probability, but many of those scenarios are extremely unlikely. 
Thus, a scenario $\bx \in \cX$ is a $\nadopt$-dimensional binary vector describing the state of adoption for every agent in the network at the end of the simulation period $T$, with $j^\text{th}$ entry given by $x_j = a_{T, j}$.

\subsection{Measuring Distribution Grid Impact}
Assessing the impact of different PV adoption scenarios requires a measure of grid stress. Under a given adoption pattern $\bx \in \scens$
solving a power flow determines the voltage at each bus and the power flow through each line in the distribution network. For distributed PV planning studies, for example, the power flow evaluation is performed at the minimum daytime load, when solar radiation and reverse power flows are highest. 

\begin{figure}[!htb]
    \centering
    \includegraphics[width=0.245\textwidth]{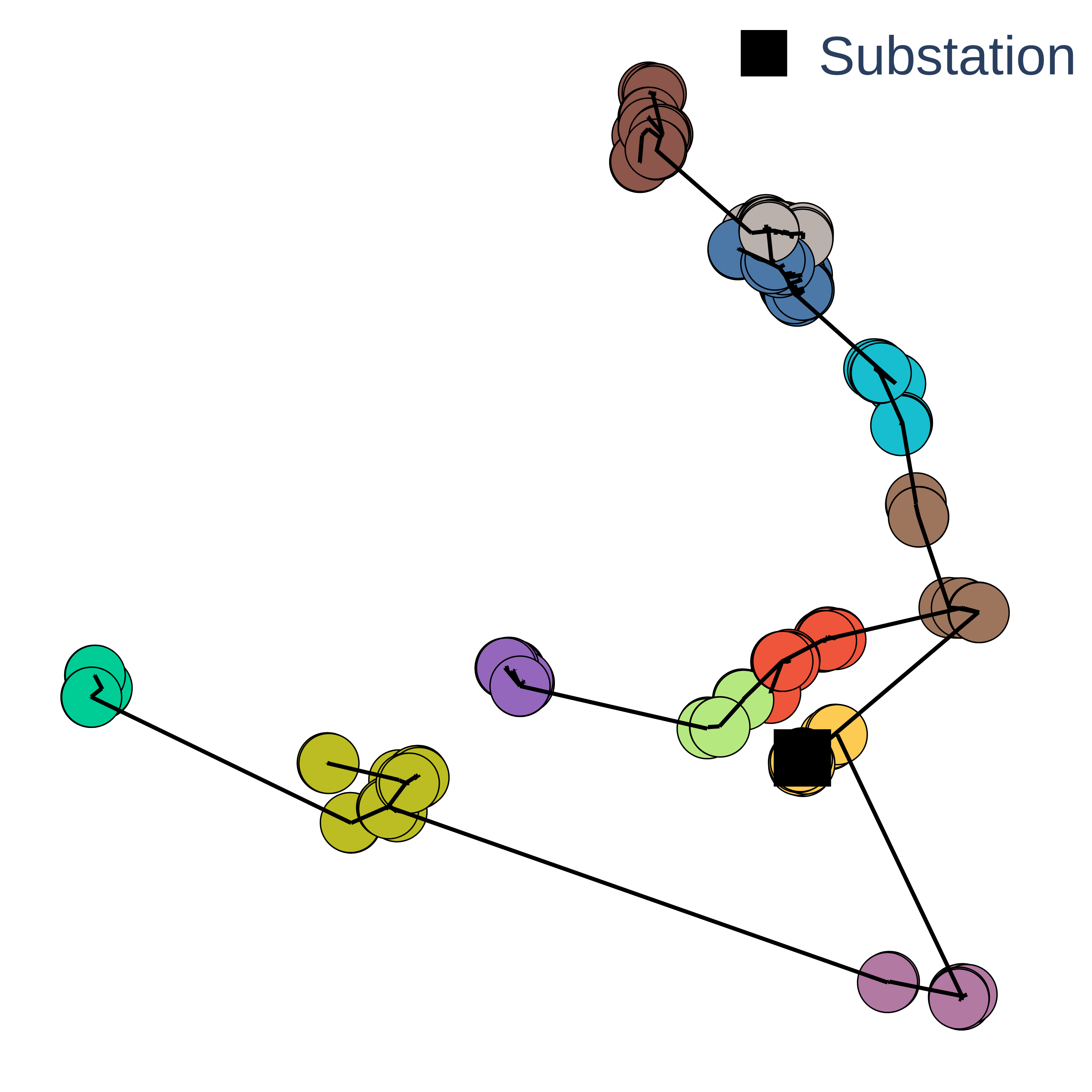}
    \hspace{-0.5cm}
    \includegraphics[width=0.245\textwidth]{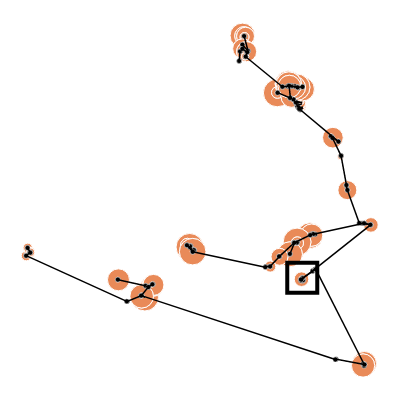}
    \caption{Feeder p4rhs8 with $248$ buses, $202$ lines and $159$ potential PV adopters, cf.~Table  \ref{tab:feeder}. \emph{Left:} partition of buses into 12 bus objectives (different colors) via the Louvain community algorithm.
    \emph{Right:} A representative critical scenario of PV adoption; circle size is proportional to installed PV capacity.}
    \label{fig:groups}
\end{figure}

We represent this grid stress as a vector-valued function $\pmb{f}$ of the scenario $\bx$ and evaluate it at both buses and lines as function of potential planning solutions. For buses, we group them into $P$ clusters $\nodes_{1},\ldots,\nodes_{P}$ that partition $\nodes$, representing voltage control zones within the distribution grid. The voltage violation in each cluster (or voltage zone) is measured by the maximum deviation of the computed bus voltage---measured per unit (p.u.) relative to the nominal system voltage---from its acceptable operating range (often $\pm5\%$). This measure quantifies the magnitude of the voltage control problem in each zone, which can be addressed by planning solutions, such as placement of capacitor banks (for undervoltage problems) or voltage regulators (for both over and undervoltage problems). The voltage zones (and number of clusters) can be defined \textit{a priori} by the utility or automatically obtained by clustering techniques. The left panel of Figure \ref{fig:groups} shows an illustrative partition of buses into 12 clusters created with the Louvain community algorithm \cite{traag_louvain_2019} for a representative feeder (p4rhs8). 


For distribution lines, stress is defined by the deviation of the power flow (measured in p.u.) from the line capacity. 

Thus, for a given scenario $\bx$, let $volt_b(\bx)$ be the computed voltage in p.u. for bus $b \in \nodes$, and let $v_b^u$, $v_b^b$ be its upper and lower voltage tolerance bounds.  Furthermore, for a line $k \in \lines := \{P + 1, \ldots, P+L\}$, let $flow_k(\bx)$ be the computed power flow and $flow^M_k$ be its capacity, i.e.~the maximum power flow line $k$ can accommodate. Then, the stress function (using common notation for buses and lines) is defined by 
\begin{equation} \label{eq:stress_bus}
f_{k}(\bx) = 
\begin{cases}
    \begin{aligned}
        & \max_{b \in \mathcal{B}_k}\left\{\max\left(volt_b(\bx) - v_b^u, v_b^l - volt_b(\bx)\right)\right\} \\
        & \qquad\qquad  \text{if } k \in \{1, \ldots, P\}
    \end{aligned}  \\ \\
    \begin{aligned}
        & flow_k(\bx) - flow^M_k \\ 
        & \qquad \qquad  \text{if } k \in \{P+1, \ldots, P+L\}.
    \end{aligned} 
\end{cases}
\end{equation}

Positive stress values $f_k(\bx) > 0$ quantify the severity of the bus or line violation. Otherwise, for a bus group, 
$f_k(\bx) \leq 0$ measures how close the most stressed bus in $\nodes_k$ is to be violated. Similarly, for a line, 
$f_k(\bx) \leq 0$ measures how close line $k$ flow is to being overloaded. 
We collect all objectives in $\pmb{f}(\bx) = \left[ f_{1}(\bx), \ldots,f_{P}(\bx), f_{P+1}(\bx),\ldots,f_{P + L}(\bx) \right]^T$. 

\subsection{Defining Critical Scenarios}

To define critical scenarios, we introduce the mapping
\begin{align} \label{eq:viol_map}
    & \Viol{y} = 
    \begin{cases} 
        y^+ & k \le P\\ 
        \sum_{j=0}^\infty j\times\mathbb{I}_{\left[ c_j, c_{j+1}\right]}(y^+) & k > P
    \end{cases}
\end{align}
where $y^+ = \max(y, 0)$. $V$ maps stress values to violation objectives. For bus objectives $k \le P$, it simply takes the magnitude of the violation in each control zone, which will determine the voltage control solutions in the investment planning. For line objectives $k>P$, $V_k$ bins excess power flows by levels of severity with respect to pre-specified thresholds $c_j$. This mirrors real-world practices, where utility planners often flag lines for reinforcement or upgrades when certain severity thresholds are exceeded (corresponding to ampacity upgrade options in reconductoring). Thus, for a given line, overloads that fall under the same $c_j$ bracket require the same reconductoring solution and are equivalent from the grid needs assessment perspective.  

To rank scenarios with violations, we introduce the dominance relation $\bx' \succ \bx$, such that $\bx'$ dominates $\bx$ if for all objectives $k$, $V_k(f_k(\bx)) \leq V_k(f_k(\bx'))$ and $V_k(f_k(\bx)) < V_k(f_k(\bx'))$ for at least one  $k$. We call a scenario $\bx$ \emph{critical} if it is in the Pareto set with respect to $V(\pmb{f})$, where the $(P+L)$-vector $V\left(\pmb{f}(\bx)\right)$ stacks all the $V_k(f_k(\bx))$. The Pareto set $\cP(\cX) = \{\bx \in \cX : \ \nexists \ \bx' \in \cX \ s.t. \ \bx' \succ \bx  \}$ is the set of all non-dominated scenarios. The image of $\cP(\cX)$ under $V(\pmb{f})$ is called the \emph{Pareto front}. 

\begin{figure}
    \centering
    \includegraphics[width=0.8\linewidth]{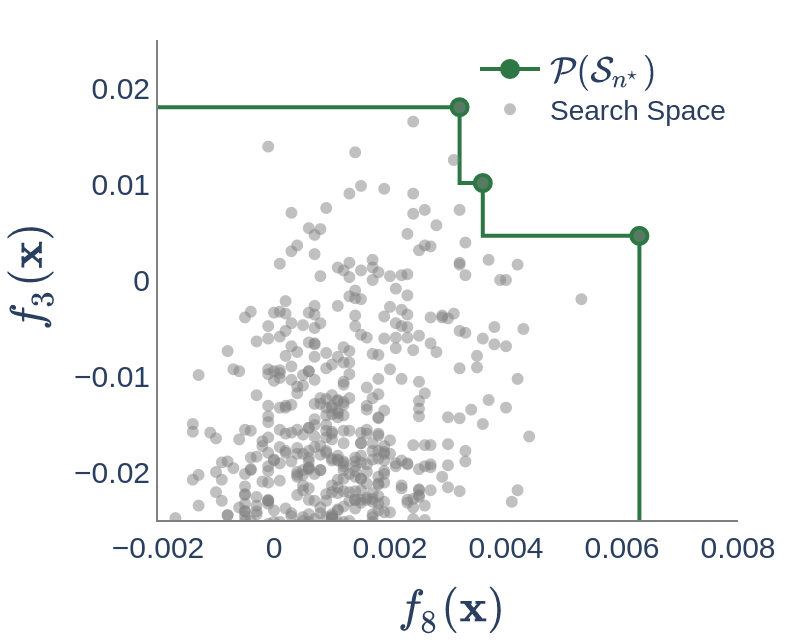}
    \caption{Pareto frontier $\cP(\cS_{n^*})$ projected on two representative bus objectives $f_3$ and $f_8$. Each point  represents the stress objective values $(f_3(\bx), f_8(\bx))$ of a scenario $\bx$. The three green points on the staircase are the critical non-dominated scenarios. }
    \label{fig:pareto_front}
\end{figure}

Using $\succ$ we map the search for critical scenarios, i.e.~$\bx$'s such that there are no other 
scenarios with a more serious violation in any bus group or line, to the task of identifying the Pareto set in the multi-objective optimization problem  
\begin{equation} \label{eq:main_problem}
	\max_{\bx\in\cX}\ V\left(\pmb{f}(\bx)\right).
\end{equation}
We call an objective $k \in \{1, \ldots, P+L\}$ critical if it can be violated due to PV adoption: $\exists \bx \in \cX: V_k({f}_k(\bx)) > 0$. Note that many objectives are typically non-critical, i.e.~never bind. The right panel of Figure \ref{fig:groups} illustrates one critical scenario for the representative feeder p4rhs8 and  Figure \ref{fig:pareto_front} illustrates the Pareto frontier with respect to two bus objectives. Each point represents a different adoption scenario $\bx$ with those lying on the staircase line being critical scenarios. 

\edit{
\subsection{Use in Distribution Grid Planning}
Given the definitions above, the set of critical DER adoption scenarios in the Pareto frontier captures the worst-case violation patterns across all possible adoption decisions. For distribution planning, this small set can replace the high-dimensional space of all adoption realizations. From a utility perspective, this greatly reduces planning complexity regardless of the methodology used. In less sophisticated environments, where planning is still manual, engineers can easily evaluate this small set of critical scenarios. In more advanced utilities, where optimal expansion planning is applied, these scenarios can serve as inputs to more robust or scenario-based optimization.}

\section{Proposed Methodology}\label{sec:bo_methodology}
To search for critical scenarios, we propose an algorithm based on BO \cite{frazier_tutorial_2018}. The main spirit of BO is to guide the search for critical scenarios using a surrogate-based acquisition function that balances exploration and \edit{exploitation}.  Rather than directly maximizing the $f_k$'s which requires extensive power flow evaluations, BO constructs a statistical model that uses previously evaluated scenarios to learn the relationship between scenarios and stress objectives. It then carries out a targeted search that only evaluates scenarios that iteratively maximize the acquisition function $\alpha: \cX \rightarrow \mathbb{R}$, approximating the probability of a scenario being critical. 

\begin{figure}[!h]
    \centering
    \includegraphics[width=\linewidth]{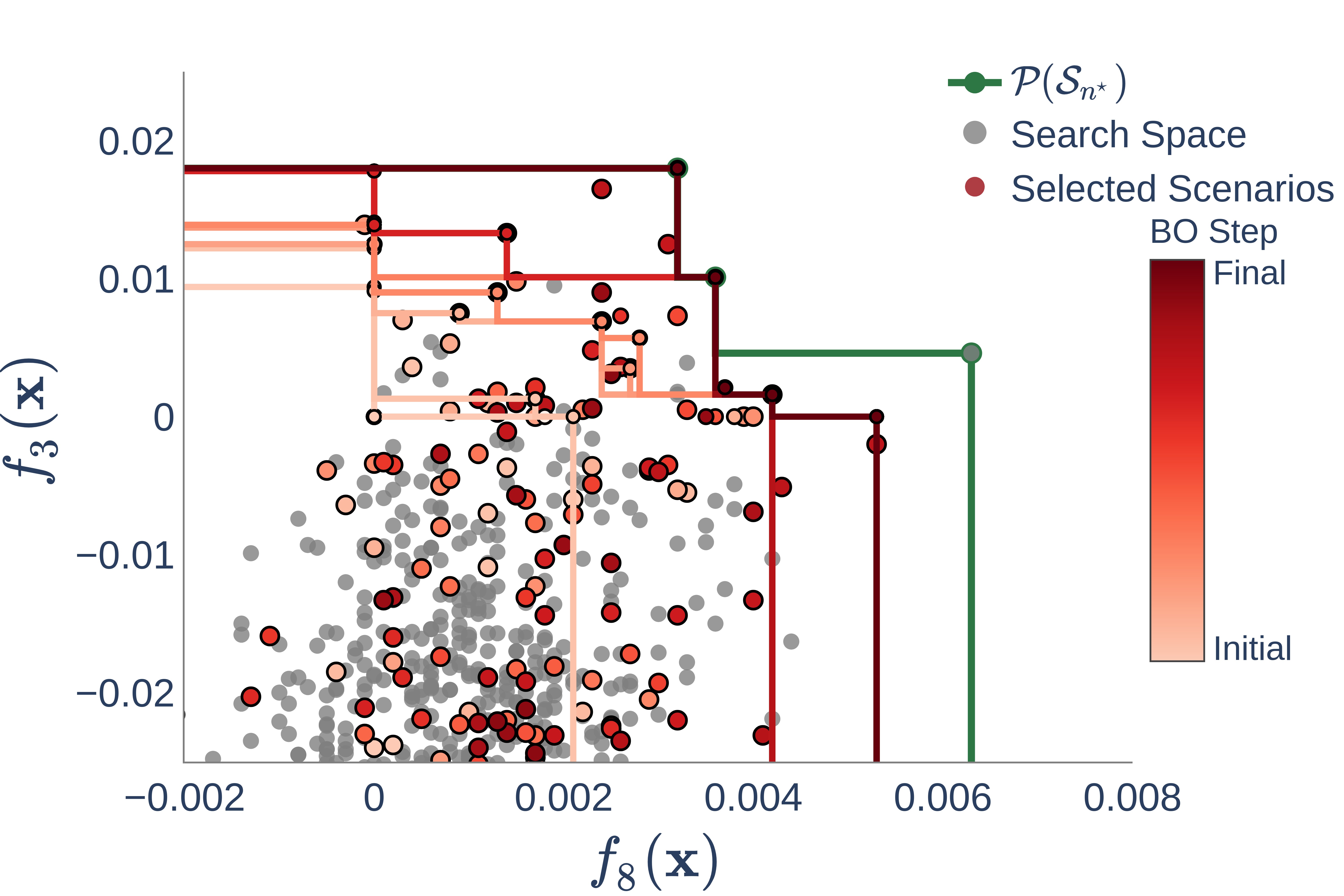}
    \caption{Progression of the Pareto frontier projected onto two representative bus objectives. Each point represents the objective values ($f_3(\bx)$, $f_8(\bx)$) of a scenario $\bx \in \cS_{\Nstop}$. Evaluated scenarios $\bx_n$ (and respective Pareto fronts $\cP_n$) are color-coded in terms of the step $n$ of the BO algorithm. \edit{Non-evaluated scenarios are shown in gray.}
    \label{fig:bo_progression-a} }
\end{figure}

The mechanics of this search 
are illustrated in Figure \ref{fig:bo_progression-a}. The BO algorithm sequentially evaluates chosen scenarios $\bx_n$ (shown by colored points) among the vast search space $\cX$. The staircase lines represent the Pareto fronts (i.e.~stresses of critical scenarios) at different steps $n$ of the search. Observe how the frontier is progressively ``pushed out''  towards the upper-right, corresponding to finding scenarios with stronger and stronger violation combos.  

\subsection{Gaussian Process Surrogates for Grid Stresses}

The statistical surrogate that we use to predict the likelihood of unevaluated scenarios being critical is a GP. GPs are analytically tractable to integrate with many acquisition functions, have good predictive performance and can naturally quantify the uncertainty of predictions.
Roughly speaking, the GP interpolates the stresses of evaluated scenarios to predict the stress of unevaluated ones. A separate GP is built for each objective $f_k$. To this end, the GP employs a probabilistic framework, constructing a covariance kernel $\kappa_k(\cdot, \cdot)$ that represents dependence among scenarios and imposing a Gaussian distribution on the predicted stress of the un-evaluated scenarios. An important challenge is handling our representation of  PV adoption scenarios as binary vectors $\bx$.

Related ideas for BO over combinatorial input spaces include
 choosing an alternative surrogate that can handle high-dim./combinatorial inputs \edit{\cite{hutter2011sequential}}, \cite{ springenberg_bayesian_2016}, performing the search in a lower-dimensional and/or continuous embedding \cite{binois_warped_2015, binois_choice_2020} 
or modifying the GP kernel to encode the structure of the objective function with respect to the input space \cite{oh_combinatorial_2019,  wan_think_2021}.

Here, we adapt the method of Wan et al.~\cite{wan_think_2021} who propose a separable categorical GP kernel based on the Hamming distance between inputs, where relevant ``active" dimensions are automatically given higher importance in predicting the output.  The categorical kernel is a product of $A$ univariate terms 
\begin{equation}\label{eq:cat_kernel}
    \kappa_k(\bx_1, \bx_2) := \eta_k \exp\left( \frac{-1}{\nadopt} \sum_{j=1}^\nadopt \theta_{k, j}\ind{x_{1, j}\neq x_{2, j}}\right).
\end{equation}

Specifically, suppose we have evaluated $n$ scenarios $\bX_n = \left[ \bx_1, \ldots, \bx_n \right]$ with corresponding stress $f_k(\bX_n) = \left[ f_k(\bx_1), \ldots, f_k(\bx_n) \right]$, $k=1, \ldots, P+L$, summarized as $\cD_n = \{\left(\bx_i, \pmb{f}(\bx_i)\right), i=1,\ldots,n\}$. Then, the stress of new, unevaluated scenarios $\bX' = \left[\bx'_1, \ldots, \bx'_M\right]$ has the predictive normal distribution  
\begin{equation} \label{eq:gpdist}
	f_k(\bX') | \cD_n \sim \mathcal{N}\left(\mu^{(n)}_k(\bX'),\Sigma^{(n)}_k(\bX')\right)
\end{equation}
where letting $\bK_k^{(n)} := K_k(\bX_n, \bX_n)$ with $\bK_{k,i,j}:= \kappa_k( \bx_i, \bx_j)$ be the GP design matrix, we have
\begin{align}\label{eq:gp_meanvar}
	\begin{split}
     \mu^{(n)}_k(\bX') &= K_k(\bX', \bX_n) [\bK_k^{(n)} + \sigma^2_k \mathbf{I}_{n}]^{-1}f_k(\bX_n); \\
	\Sigma^{(n)}_k(\bX') &= K_k(\bX', \bX') \\ 
    - &K_k(\bX', \bX_n) [\bK_k^{(n)} + \sigma^2_k \mathbf{I}_n]^{-1} K_k(\bX_n, \bX').
	\end{split}
\end{align}
In \eqref{eq:gp_meanvar}
 $K_k(\bX_n, \bX')$ is the \textit{Gram} matrix containing pairwise correlations between considered scenarios, and $\mathbf{I}_n$ the $n\times n$ identity matrix. The observation variance $\sigma_k^2$ is a nugget term, which 
ensures numerical stability during the modeling process. The hyperparameters $\theta_{k, j}$ in \eqref{eq:cat_kernel} are inferred using maximum likelihood estimation based on $\cD_n$, which allows the GP to learn the adopters $j$ that most affect each objective $f_k$, a process referred to as Automatic Relevance Determination (ARD). 
The posterior mean $\mu_k^{(n)}(\bx')$ represents the predicted stress of objective $k$ at scenario $\bx'$ and the posterior standard deviation $\sigma^{(n)}_k(\bx')$ is used for uncertainty quantification to measure the confidence in the above prediction, employed below to enforce exploration in the scenario space.

\subsection{Acquisition Function}\label{sec:acqf}
\edit{In order to guide the search for scenarios on the Pareto front, we construct
an acquisition function $\mathcal{\alpha}_{ND}(\cdot)$ (where ``ND" stands for non-dominated) that proxies the probability of scenario $\bx$ being non-dominated with respect to $V(\pmb{f})$. Thus, the idea is to have $\alpha_{ND}(\bx; \cD_n) \approx \proba{\bx \in \cP(\cX) | \cD_n}$, so that maximizing $\alpha_{ND}$ corresponds to proposing scenarios that are most likely to 
improve the current Pareto front $\cP(\bX_n)$.}

\edit{Because it is important to incorporate the covariance structure among unobserved scenarios and rank their potential violations consistently, we evaluate  $\mathcal{\alpha}_{ND}$ in parallel over $M$ candidate scenarios $\bX'_n = \begin{bmatrix}\bx'_1, \ldots, \bx'_M \end{bmatrix}$.  To do so, we first draw $N$ samples of $\pmb{f}(\bX'_n) | \cD_n$ using the latest GPs, denoted as $\widetilde{\pmb{f}}_1(\bX'_n), \ldots, \widetilde{\pmb{f}}_N(\bX'_n)$. These capture the distribution of potential violations forecasted for $\bX'_n$. We then compute the subset of critical scenarios $\widetilde{\cP}_i \subseteq \bX'_n$ in each of these $N$ samples. Then for a candidate $\bx_m \in \bX'_n$, we set }
\begin{equation}\label{eq:prob_crit}
    \alpha_{ND}(\bx_m; \cD_n) := \frac{1}{N}\sum_{i=1}^N \mathbb{I}_{\widetilde{\cP}_i}(\bx_m). 
\end{equation}
\edit{
Thus, $\alpha_{ND}(\bx_m)$ is the frequency that $\bx_m$ is critical among $\bX'_n \cup \bX_n$.} 

\begin{figure}
    \centering
    \includegraphics[width=\linewidth]{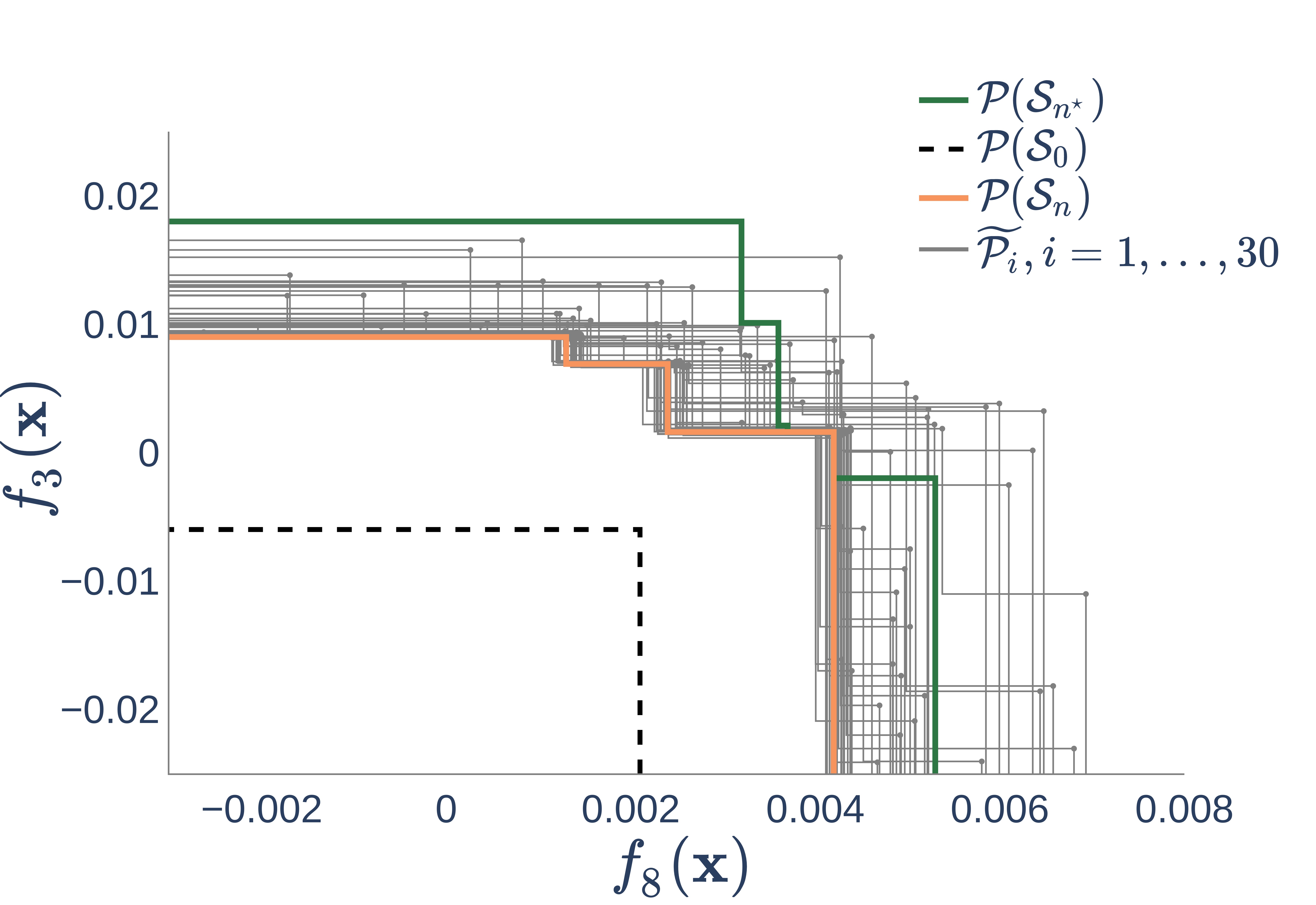}
    \caption{\edit{Initial Pareto front $\cP(\cS_0)$ (dashed gray), Pareto front $\cP(\cS_{n})$ at step n = 55 (orange) and true front $\cP(\cS_{\Nstop})$ (green) across two representative objectives $f_3(\bx)$, $f_8(\bx)$. We also plot 30 simulated fronts $\widetilde{\cP}_i$ (solid gray curves) obtained by sampling the terminal GP across $M=500$ non-evaluated scenarios.} 
    \label{fig:bo_progression-b}}
\end{figure}

\edit{The conditional simulation of Pareto fronts closely follows the approach of \cite{binois_quantifying_2015}, although here we map simulated samples to violations. This construction  is visualized in Figure \ref{fig:bo_progression-b} which shows 30 simulated Pareto fronts $\widetilde{\cP}_i$ (gray ``staircases'') for two bus objectives %
$f_1$ and $f_8$. Then, $\alpha(\bx_m; \cD_n)$ is proportional to the number of times the objectives sampled at $\bx_m$ lie on the staircase. Note that $\bx_m$ itself is not being plotted, so visually the same scenario would yield $N$ different corners.
Although a larger number of samples $N$ in \eqref{eq:prob_crit} better accounts for the covariance structure $\Sigma^{(n)}(\bX'_n)$,  leading to more accurate $\alpha_{ND}$ (and hence more accurate search) it comes at the cost of a  higher computational burden. Similarly, jointly sampling from a GP at $M$ non-evaluated scenarios $\bX'_n$ to compute $\alpha_{ND}$ and $\tau_n$ becomes prohibitive as $M$ grows. }

\subsection{Search space and stopping criterion}
To initialize the GPs and the search we begin by simulating and evaluating $n_0$ scenarios. Thereafter, in order to reflect the likelihood of scenarios under the agent-based diffusion model (\ref{eq:bass_prob}) we do not maximize the acquisition function over the entire $\cX$ but over smaller search spaces $\cS_n$ that grow in $n$. To this end, we first  
simulate $N_\text{init}$ scenarios to construct $\cS_1$ and then augment with $N_\text{expand}$ additional new scenarios at each step, so that $|\cS_n| = n_0 + N_\text{init} + (n-1) \cdot N_\text{expand}$. 


Evaluation of $\alpha_{ND}(\cdot)$ on every scenario in the search space $\cS_n$
becomes expensive when $|\cS_n| \gg 10^3$. Therefore, we further restrict to a set of $M$ candidate scenarios $\bX'_n$, where $M$ is set to the size of the initial search space $\cS_1$. As $\cS_n$ grows, we mitigate the risk of promising scenarios being overlooked by sampling them with probability inversely proportional to the frequency with which they have already has been sampled. Thus, scenarios that have never appeared in $\{\bX'_{k}, k \le n\}$ have higher probability of being sampled into $\bX'_{n+1}$.

The acquisition function $\alpha_{ND}$ is further used as a stopping criterion to terminate the search, interpreted as a statistical guarantee on the set of critical scenarios discovered, which is of particular interest to utility planners. Indeed, the probability of having missed a critical scenario can be approximated as
\begin{align}\label{eq:approx_prob}
    \begin{split}
        \proba{\exists \ \bx \in  \cP(\cX)\setminus \bX_n} & \lessapprox \sum_{\bx \in \cS_n \setminus \bX_n} \proba{\bx \in \cP(\cX)} \\
        &\approx \textstyle\sum_{\bx \in \cS_n \setminus \bX_n} \alpha_{ND}(\bx; \cD_n).
    \end{split}
\end{align}
Therefore, the search is stopped when this upper bound has crossed a specified tolerance $\bar{\tau}$, set by the user. Since $\cS_n$ grows at every step $n$, we compute \eqref{eq:approx_prob} on a random subsample of $M$ unevaluated scenarios $\bX' \in \cS_n \setminus \bX'_n$
\begin{equation}\label{eq:stop_crit}
    \tau_n := \textstyle\sum_{\bx' \in \bX'} \alpha_{ND}(\bx'; \cD_n).
\end{equation}
Our methodology is summarized in Algorithm \ref{alg:dumbo}.

\begin{algorithm}[!h]
\caption{Critical Scenario Search with Bayesian Optimization}
\label{alg:dumbo}
\begin{algorithmic}[1]
\State Initialize $\cD_0 = \{(\bX_0, \pmb{f}(\bX_0))\}$ w/ $n_0$ evaluated scenarios
\State Set $n \leftarrow 0$ and $\tau_n \leftarrow \infty$
\While{$\tau_n > \bar{\tau} $} \hspace{1.87in} 
\Statex \codecomment{// Search until stopping criterion is satisfied, see \eqref{eq:stop_crit}}
\State $\cC_{n} \leftarrow \{ k: \exists\bx \in \cD_{n}, f_k(\bx) > 0\}$ 
\codecomment{// Critical objectives}
\State $\mathcal{E}_n \leftarrow \{k: \max_{\bx \in \cS_n}f_k(\bx) > \bar{s}\}$
\State $\cO_n \leftarrow \cC_n \cup \ \mathcal{E}_n \qquad\qquad\qquad\quad $ \codecomment{// Active objectives}
\State Update/fit the surrogates $\mathcal{GP}( \mu_k^{(n)},  \sigma_k^{(n)})$ $\forall k \in \cO_n$ 
based on $\cD_n$
\State $\bx^{(n)} \leftarrow \argmax_{\bx \in \cS_n }\alpha_{ND}(\bx; \cD_n) \qquad\quad$  \codecomment{// see \eqref{eq:prob_crit}}
\Statex \codecomment{// Evaluate and augment to training dataset }
\State Evaluate objectives $\pmb{f}(\bx^{(n)})$
\State $\cD_{n} \leftarrow \cD_{n-1} \cup \{(\bx^{(n)}, \pmb{f}(\bx^{(n)})\}$ 
\Statex \codecomment{// New scenarios to test}
\State $\cS_n \leftarrow \cS_n \cup \{ N_\text{expand} \text{ fresh scenarios  from }\cX\}$ 
\State $n \leftarrow n+1$ 

\EndWhile
\State Return violating scenarios $\{\bx \in \bX_{\Nstop}: \exists k \ f_k(\bx) > 0 \}$, found critical objectives $\cC_{\Nstop}$ and critical scenarios $\cP(\cS_{\Nstop})$
\end{algorithmic}
\end{algorithm}

In order to speed up the search, we make a few adjustments to Algorithm \ref{alg:dumbo}. Firstly, there is no need to model objectives that have no violations as they do not affect the Pareto front, i.e.~it is sufficient to focus on critical objectives only. 
%
%
Since the latter are not known until we have evaluated a scenario that leads them to be violated, we compute $\alpha_{ND}$ based on a set $\cO_n = \cC_n \cup \mathcal{E}_n$ (line 7) where $\cC_n$ is the set of objectives that have already been violated, and  $\cE_n$ is the set of ``stressed" objectives, i.e.~those that are close to being violated. We take $\mathcal{E}_n = \{k: \max_{\bx \in \cS_n}f_k(\bx) > \bar{s}\}$ where $\bar{s} <0$ is a threshold decided by the user. The idea is to look for buses and lines that are operating close to their tolerance/capacity, and hence might actually be violated in yet-to-be evaluated scenarios. 
Thus, including such objectives in the acquisition function will guide the search towards scenarios that are likely to push the objective towards a violation, thereby ``detecting" a new critical objective. Thus, when computing $\alpha_{ND}$, the Pareto fronts $\widetilde{\cP}_1,\ldots, \widetilde{\cP}_N$ are taken with respect to objectives $V_k(\pmb{f}_k)$, for $k\in\cO_n$. Furthermore, we alternately let active objectives $\cO_n$ be either bus or line objectives, but not both simultaneously. To be consistent with this choice, we track a stopping criterion for both buses and lines, denoted by $\tau_n^b$ and $\tau_n^l$, and terminate the search when $\tau_n := \max(\tau_n^b, \tau_n^l) < \bar{\tau}$. 

Secondly, fitting GP hyperparameters at every iteration via minimization of the log-likelihood (line 8) unnecessarily increases the computational burden without yielding significant improvements in model performance. GP hyperparameters tend to stabilize after a few iterations and minor changes to them have minimal impact on the BO search. Therefore, we refit the GP hyperparameters (while still updating the posterior means/standard deviations as new evaluations are added) only every $N_0$ steps.
%
%
%
Finally, to reduce overhead, we select the top $B=4$ scenarios (rather than just the top-1 in \edit{line 8)} that maximize $\alpha_{ND}$, evaluating this batch in parallel. 


\section{Numerical Results} \label{sec:results}

We implement our  approach in Python, using the GP implementation from the GPyTorch\footnote{https://gpytorch.ai} and BoTorch\footnote{https://botorch.org} open source libraries \cite{balandat_botorch_2020}. We built our own acquisition functions and kernels on top of the modules available in BoTorch. 

We proceed to a case study applying Algorithm \ref{alg:dumbo} to identify critical grid stress scenarios on the real-life feeder, named \textit{p4rhs8} with 12 bus groups shown in Figure \ref{fig:groups}. Next we will analyze a larger feeder, \textit{p3uhs27}, to show consistency of the methodology proposed. The networks are obtained from National Renewable Energy Laboratory's SMART-DS open dataset \footnote{https://data.openei.org/submissions/2981} with the same names as in the paper. PV capacities of adopting agents are computed using the PySAM\footnote{https://sam.nrel.gov/software-development-kit-sdk/pysam.html} library. Bus voltage and line flows are computed using an optimal power flow solver with the pandapower\footnote{https://www.pandapower.org} open source library for Python.


\subsection{Algorithm Performance} \label{sec:feeder-1}
\begin{table}[h!]
    \centering
    \begin{tabular}{p{3.5cm}rr|rr}
    & \multicolumn{2}{c|}{\textbf{p4rhs8}} & \multicolumn{2}{c}{\textbf{p3uhs27}}\\
    \hline
       \multicolumn{5}{c}{\textbf{Feeder Characteristics}} \\ 
    \hline
    Number of buses & \multicolumn{2}{r|}{202} & \multicolumn{2}{r}{439} \\
    Number of lines & \multicolumn{2}{r|}{202 (29 miles)} & \multicolumn{2}{r}{339 (13 miles)} \\
    Peak load demand (MW) & \multicolumn{2}{r|}{0.26} & \multicolumn{2}{r}{1.08} \\
    Energy Consumption (GWh) & \multicolumn{2}{r|}{6.37} & \multicolumn{2}{r}{15.5} \\
    Number of agents ($A$) & \multicolumn{2}{r|}{159} & \multicolumn{2}{r}{329}\\
    Voltage tolerance $(v_l^b, v_b^b)$ & \multicolumn{2}{r|}{(0.95, 1.05)} & \multicolumn{2}{r}{(0.965, 1.035)}\\
    \hline
     \multicolumn{5}{c}{\textbf{Scenario / Algorithm Settings} } \\ 
    \hline
    Initial search space $|\cS_\text{1}|$ & \multicolumn{2}{r|}{3000} & \multicolumn{2}{r}{3000} \\  
    Initial sample size $n_0$ & \multicolumn{2}{r|}{75} & \multicolumn{2}{r}{100}\\  
    Simulations per step $N_\text{expand}$ & \multicolumn{2}{r|}{200} & \multicolumn{2}{r}{300}\\  
    Simulated scenarios $|\cS_{\Nstop}|$ & \multicolumn{2}{r|}{9200} & \multicolumn{2}{r}{21800}\\
    Evaluated scenarios $|\bX_{\Nstop}|$ & \multicolumn{2}{r|}{333} & \multicolumn{2}{r}{452} \\
    Stopping threshold $\bar{\tau}$ & \multicolumn{2}{r|}{0.1} & \multicolumn{2}{r}{0.2} \\
    Evaluation batch size $B$  & \multicolumn{2}{r|}{4} & \multicolumn{2}{r}{5} \\
    \hline
     \multicolumn{5}{c}{\textbf{Results} } \\ 
    \hline
    & Buses & Lines & Buses & Lines \\
    Objectives & 12 & 202 & 9 & 339 \\
    Violating scenarios found & 213 & 255 & 337 & 281 \\
    Percentage found & 82.56\% & 98.84\% & 95.74\% & 79.83\% \\
    Crit. objectives $|\cC^*_{\Nstop}|$ & 5 & 10 & 4 & 6 \\
    Crit. objectives found $|\cC_{\Nstop}|$ & 5 & 10 & 4 & 6\\
    Crit. scenarios $|\cP(\cS_{\Nstop})|$ & 15 & 32 & 36 & 20\\
    Crit. scenarios found $|\cP(\cS_{\Nstop}) \cap X_{\Nstop}|$ & 13 & 31 & 34 & 19 \\
    \hline\hline
    \end{tabular}
    \caption{Summary of settings and search results for feeders p4rhs8 and p3uhs27. The number of violating scenarios discovered is the size of the set $\{\bx \in \bX_{\Nstop}: f_k(\bx) > 0 \text{ for some } k\}$.}
    \label{tab:feeder}
\end{table}

The settings and main results are summarized in Table \ref{tab:feeder}. All scenarios are generated from the agent diffusion model \eqref{eq:bass_prob}  with $p=0.01$ and $q=0.164$. We first simulate $n_0=75$ scenarios to evaluate power flows and initialize the GPs. Algorithm \ref{alg:dumbo} then iteratively expands the scenario space, simulating $N_\text{expand} = 200$ scenarios per iteration, except at the first step where we simulate an extra $2800$ scenarios (thus $\cS_1 = 3000)$. The stopping criterion \eqref{eq:stop_crit} with $\bar{\tau}=0.1$ is used to automatically terminate the search.
The shown run terminates after evaluating a total of $333$ (75 initial + 258 selected) scenarios. 

Figure \ref{fig:bo_progression-a} shows the progression of the Pareto frontier projected on two of the 5 critical bus objectives. 
For better visibility, we crop part of the search space where the objectives are  negative. The ultimate goal of our methodology is to recover the critical scenarios lying on the outermost staircase line. Critical scenarios are discovered in the later stages of the BO, suggesting that it does not stop prematurely. Consistently with this fact, a few later-stage evaluations are relatively far from the frontier, probably because the algorithm focuses on the corners of the Pareto front for other critical objectives as well (where the two objectives shown might not be simultaneously violated). 
The picture looks similar for other objectives, with all but one scenario on the Pareto frontier being discovered.

\begin{figure}[!htb]
    \centering
    \includegraphics[width=0.98\linewidth]{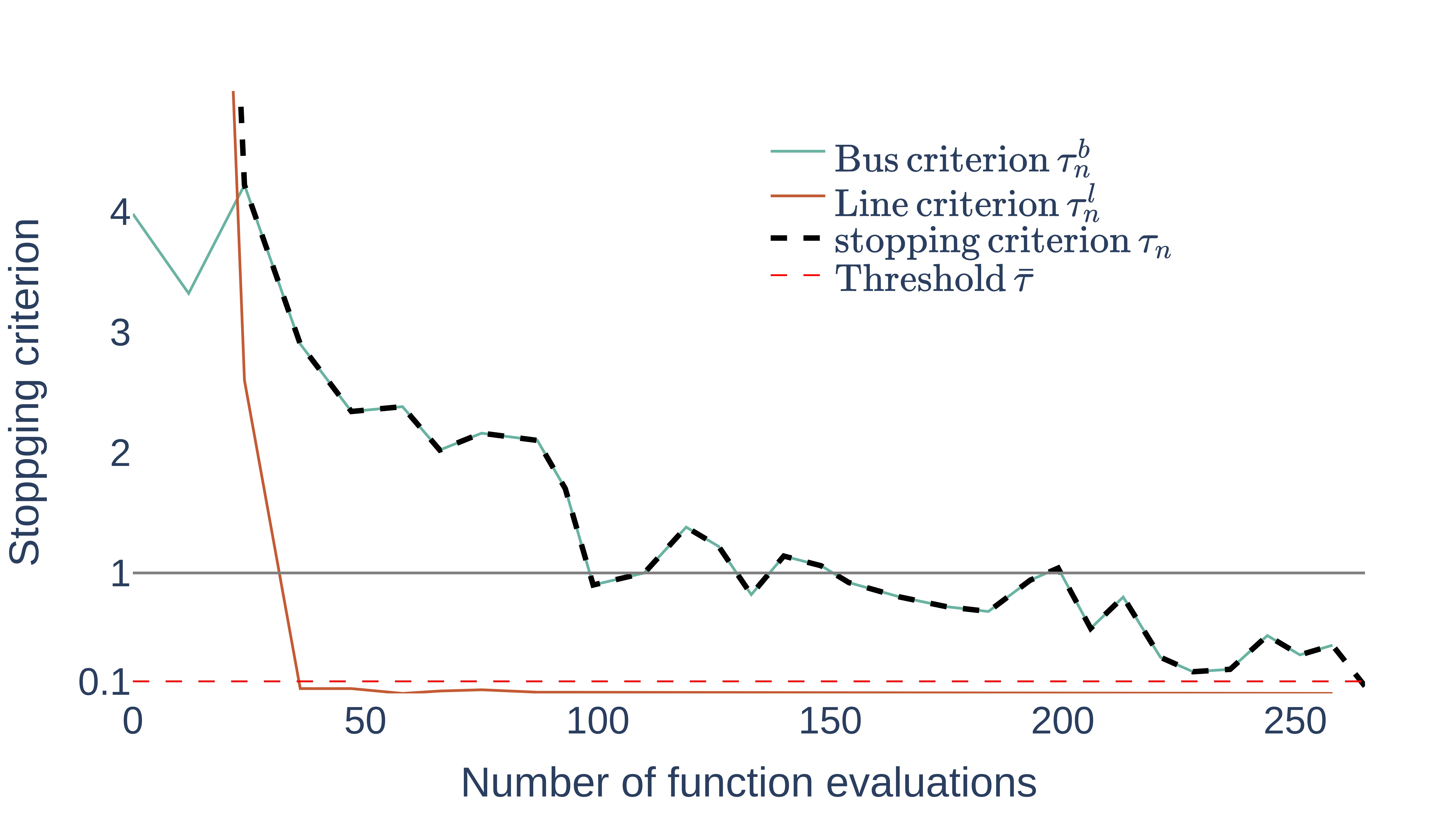}
    \caption{\edit{Evolution of statistical stopping criteria $\tau_n^b$ for buses and $\tau_n^l$ for lines as a function of number of evaluations $n$. The overall stopping criterion  $\tau_n = \max(\tau^b_n, \tau^l_n)$ is triggered at the threshold $\bar{\tau}=0.1$.}} 
    \label{fig:stopping_criterion}
\end{figure}

In all, we find 13 out of 15 bus-critical and 31 out of 32 line-critical scenarios among a vast set of 9200 simulated scenarios, with only 333 evaluations. The algorithm also discovers all critical objectives (5 for the buses and 10 for the lines). To get a sense of how targeted the search is, among the 333 scenarios that the BO evaluates, the vast majority cause violations during PF: there is some bus violation in over 83\% and a line violation in over 98\% of evaluated scenarios (Table~\ref{tab:feeder}).
Figure \ref{fig:stopping_criterion} shows the stopping criterion $\tau_n$ against $n$. The stopping criterion bounds from above the likelihood of missing a critical scenario, effectively stating that discovered critical scenarios account for no less than $(1-\bar{\tau})\times 100\%$ of all simulated critical scenarios. One can set $\bar{\tau}$ higher to quicken the search at the cost of lower confidence in omitting critical scenarios.

\subsection{Comparators}

One baseline competitor is a brute force approach that sequentially evaluates every simulated scenario to identify the critical ones. In the above case study, this would require 27.6$\times$ more evaluations (9200/333) and would take 7.65$\times$ longer runtime after accounting for the overhead of our algorithm including the computation of the acquisition function and fitting GPs on all the bus and line objectives. Furthermore, the brute force approach lacks statistical analysis of criticality. 

\begin{figure}
    \centering
    \includegraphics[width=\linewidth]{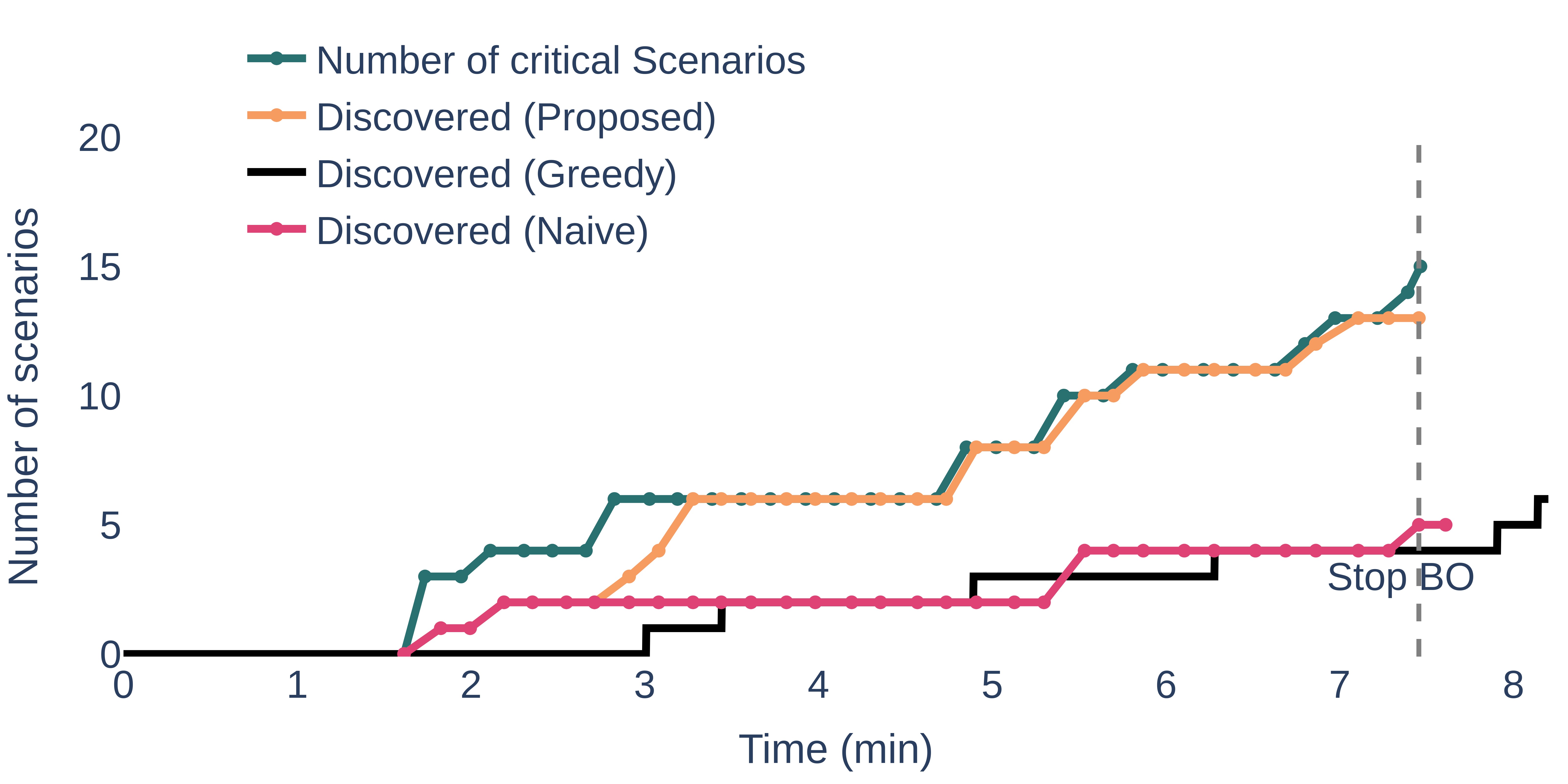}
    \caption{Number of bus-critical scenarios in  the search space $\cS_n$(green line) and how many of those scenarios are identified by Algorithm \ref{alg:dumbo} \edit{(orange line) and a greedy comparator (in black) as a function of time. The number of critical scenarios found by a naïve comparator that evaluates the $N_{top}=25$ scenarios by total PV is the magenta curve. When Algorithm \ref{alg:dumbo} stops after 7.5 minutes, the greedy approach has only identified 4 critical scenarios.}}
    \label{fig:dumbovsbf}
\end{figure}

Figure \ref{fig:dumbovsbf} illustrates how our methodology picks up bus-critical scenarios. The green curve counts the total number of critical scenarios in the search space $\cS_n$, which increases in $n$. The orange curve shows the size of the subset of these that have been selected and evaluated by the BO, i.e.~scenarios that are correctly identified as critical. Our algorithm is able to quickly identify critical scenarios shortly after they are added to the search space, cf.~the fact that the orange curve closely tracks the green one. In contrast, the black staircase representing the number of critical scenarios found by the brute force approach lags significantly behind, unable to keep up with the growing size of $\cS_n$, leading to a widening gap between the number of critical scenarios selected by our algorithm and those discovered through brute force. 

\begin{figure}[!h]
    \centering
    \includegraphics[width=\linewidth]{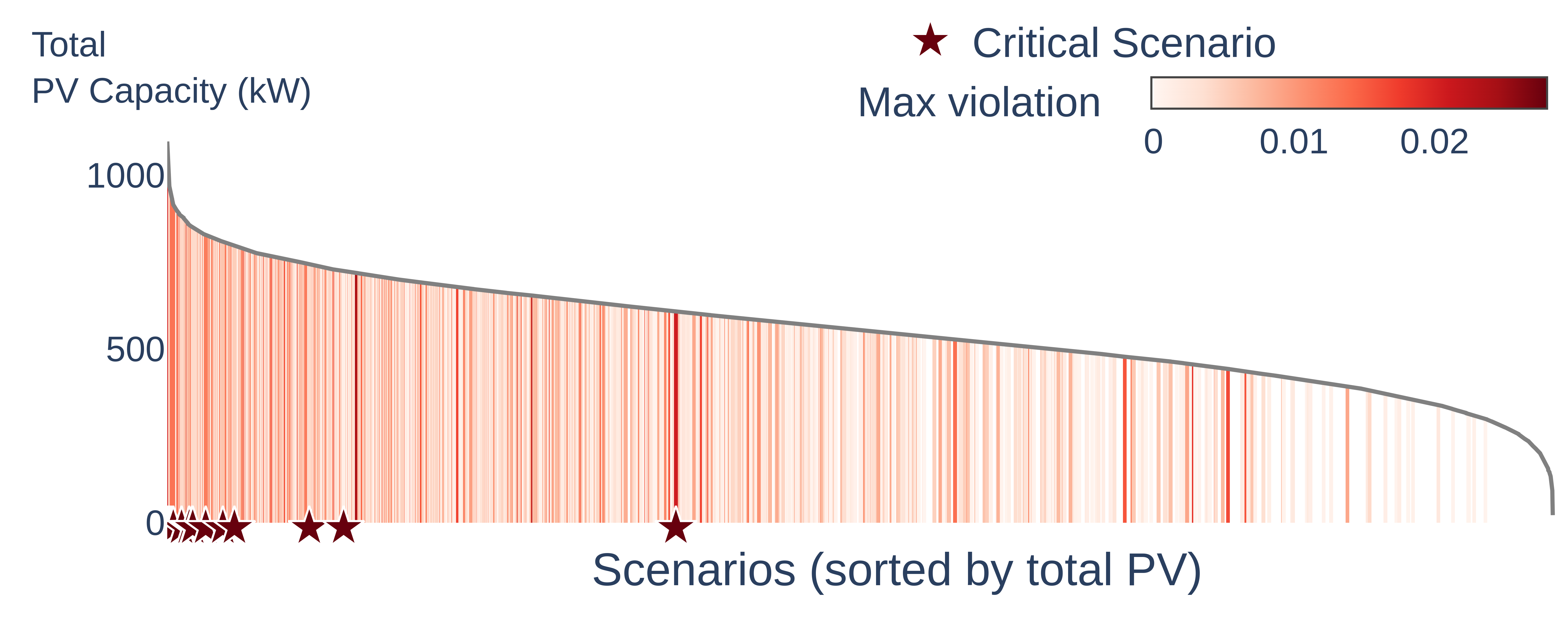}
    \includegraphics[width=\linewidth]{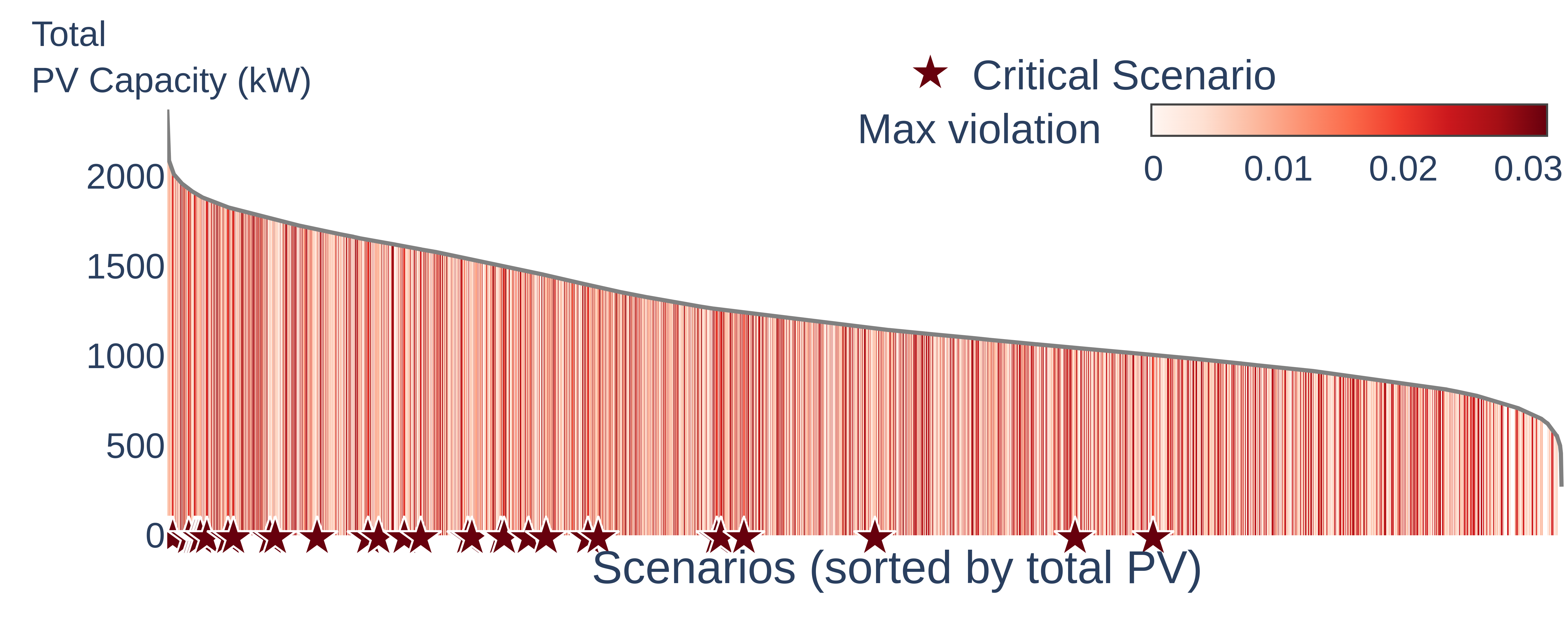}
    \caption{\edit{Bus violations within the simulated PV adoption scenarios for feeders p4rhs8 (top, 9200 scenarios) and p3uhs27 (bottom, 21800 scenarios), sorted left-to-right by their total installed PV capacity. Each vertical bar corresponds to a scenario, with color indicating the maximum bus voltage violation observed under that scenario. Adoption scenarios identified as critical by Algorithm \ref{alg:dumbo} (i.e., belonging to the Pareto set of violations) are marked with a star.}}
    \label{fig:critical_scenarios}
\end{figure}

A second comparator approach is to rank simulated scenarios by total PV adoption \edit{ and compare the critical scenarios obtained by the proposed search algorithm with the existing utility practice, which consists of selecting} the extreme cases of aggregated capacity (``high" and ``low" load or PV scenarios) as worst-cases to explore. This naïve approach evaluates the $N_{top}$ scenarios that have the highest adoption capacity. 
\edit{Figure \ref{fig:dumbovsbf} shows that this naïve approach (with $N_{top} = 25$)  misses the majority of critical scenarios that our algorithm finds and barely improves upon greedily evaluating every simulated scenario.} The top panel of Figure \ref{fig:critical_scenarios} further shows the distribution of the 9200 simulated scenarios sorted by their aggregated adopted PV, and colored by the maximum violation over the buses. \edit{Notably, most scenarios with the largest PV adoption are not critical, while other scenarios with relatively low PV capacity are critical (indicated with a star) due to specific locational adoption patterns. For example, high PV penetration throughout a feeder can lead to overvoltages at nodes distant from the substation while alleviating overloading in lines closer to it. However, a moderate PV penetration at the end of the feeder can still cause the same overvoltages but may not be sufficient to alleviate overloading, resulting in a more challenging planning scenario. This highlights the importance of systematic locational search algorithms and explains how, with our method, we successfully find critical scenarios that the greedy and naive approaches miss.} 

\begin{figure}[!h]
    \centering
    \includegraphics[width=\linewidth]{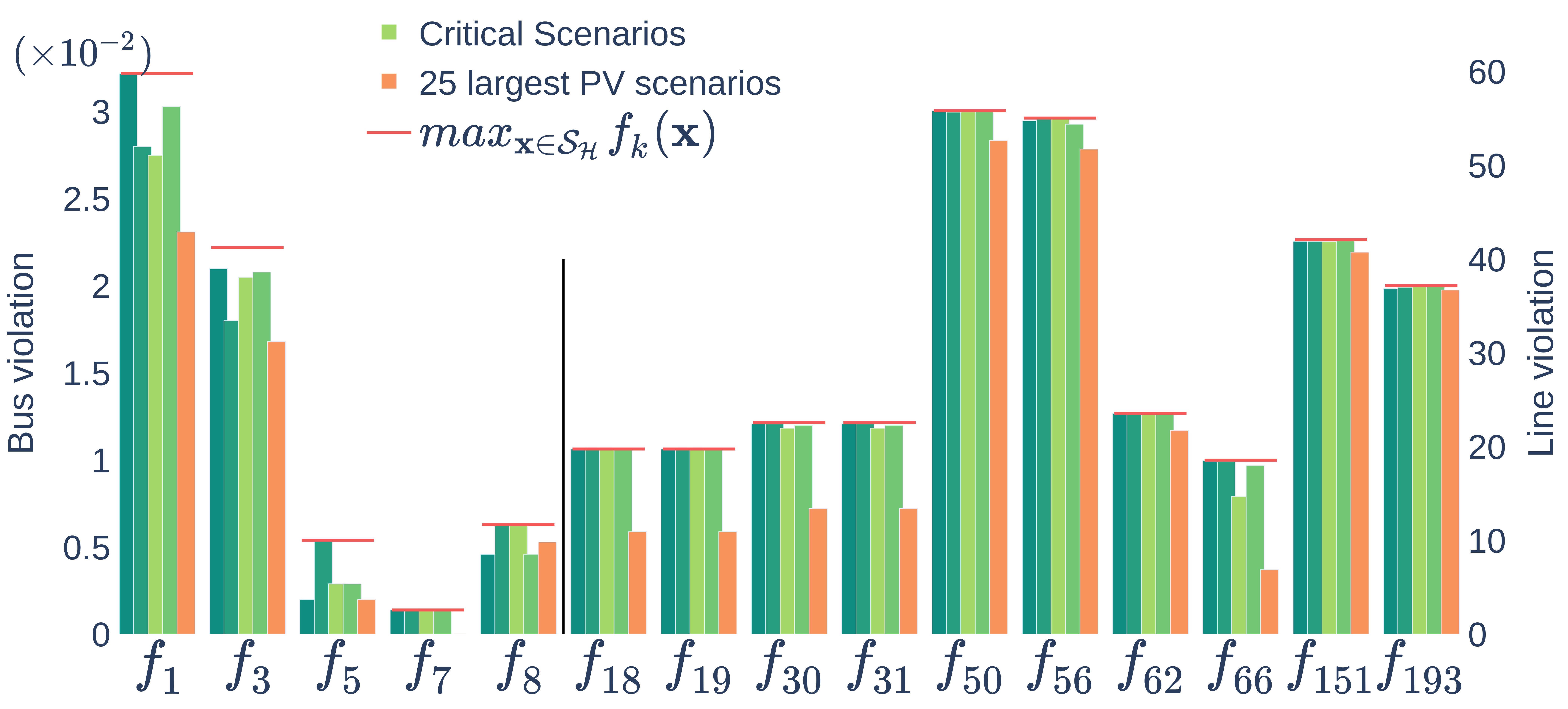}
    \caption{Critical objective values found across 4 runs (green bars) of Algorithm \ref{alg:dumbo} and a fixed database $\cS_H$ with $H=25000$ scenarios. Orange bars show the maximum violation attained in the top $N_{top}=25$ scenarios by total PV. The horizontal lines show the maximum violation \edit{in this database}. The critical bus objectives are $f_k$ (left $y$-axis) $k\in \{1, 3, 5, 7, 8\}$, and the critical line objectives (right $y$-axis) are $k\in \{18, 19, 30, 31, 50, 56, 66, 151, 193\}$.}
    \label{fig:critical_objectives}
\end{figure}

Figure \ref{fig:critical_objectives} visualizes the critical violations found versus the ground truth (orange bars) and versus the violations within the top-25 scenarios ranked by total PV capacity.
%
We see that critical scenarios of PV adoption lead to more serious violations in nearly every critical objective than the largest PV scenarios. For the bus objective $f_7$, simply looking at the largest PV scenarios would entirely miss the violation.

\subsection{Second Feeder}

\begin{figure}[!h]
    \centering
    \includegraphics[width=0.48\linewidth]{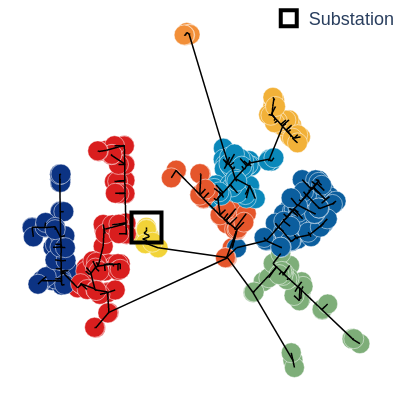}
    \hspace{-0.5cm}
    \includegraphics[width=0.52\linewidth]{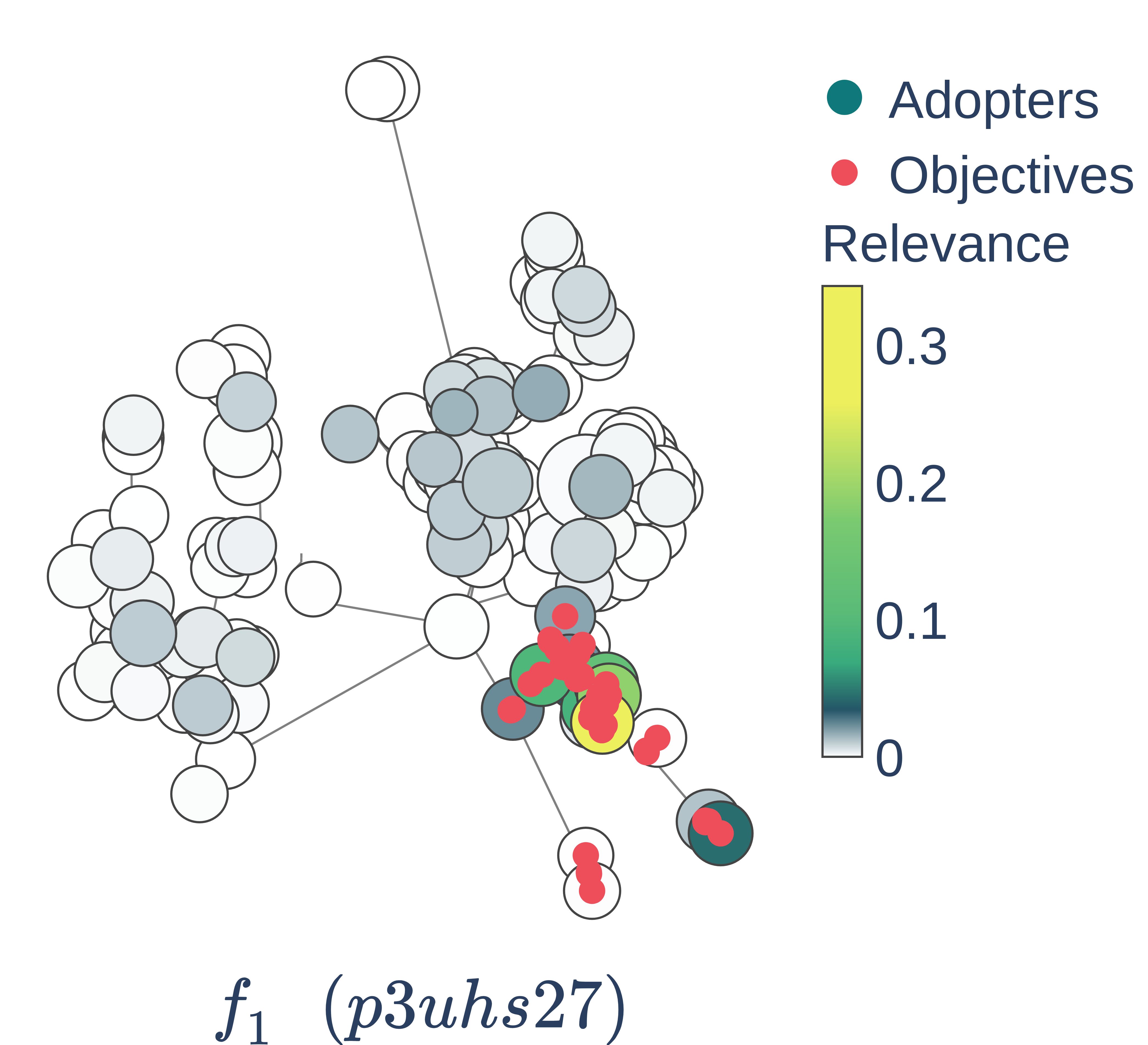}
    \caption{\emph{Left:} Feeder p3uhs27 and bus groupings obtained using the Louvain community algorithm. \emph{Right:}
    Spatial map of adopter relevance $\tilde{\theta}_{1, j}$ for a representative bus objective $f_1$. \edit{Adopters $j$ marked with an orange dot are buses that are part of objective $f_1$}. Larger values (brighter colors) indicate stronger influence of the adopter on the respective objective. Circle size is proportional to the adopter's PV system capacity.}
    \label{fig:p3uhs27}
\end{figure}

To validate the performance of our algorithm, we consider a larger 439-bus network (feeder p3uhs27) with 9 bus groups and 300+ adopters (shown in left panel of Figure \ref{fig:p3uhs27}), see the right column of Table \ref{tab:feeder}. For this feeder, a stopping threshold of $\bar{\tau} = 0.2$ is set and we use batch evaluations with $B = 5$ scenarios. 
Results are similar to the first feeder p4rhs8. The search strongly targets scenarios leading to violations (over $95\%$ of evaluated scenarios result in bus voltage violations, and over $79\%$ in line flow ones). We recover 34 out of 36 bus-critical and 19 out of 20 line-critical scenarios. This shows the robustness of our methodology. The lower panel of Figure \ref{fig:critical_scenarios} shows the distribution \edit{of the total adopted PV of the 21800 simulated scenarios and color-coded by the maximum bus violation}. The Figure highlights that violations are common (even for scenarios with lower total adopted PV capacity), but only a few scenarios are critical, and those are not necessarily the ones with the largest PV. Even if we evaluated the top 452 scenarios by total PV, we would miss a majority of the critical scenarios. 

\subsection{Algorithm Stability}\label{sec:stability}
To evaluate the stability of our search, we first fix a large database $\cS_H$ of $H = 25000$ scenarios and run Algorithm~\ref{alg:dumbo} four times on feeder p4rhs8 where for each run $i=1,\ldots,4$ the search space $\cS^{(i)}_n$ is constructed by randomly sampling without replacement from $\cS_H$, with a different seed for every run, \edit{while keeping algorithm parameters fixed}. 
For each run, we record in Table \ref{tab:algo_stability} the number of function evaluations $\Nstop$, the search space size at termination ($\cS^{(i)}_n$), and the number of found bus- and line-critical scenarios. The maximum violation achieved by evaluated scenarios (i.e.~the marginal Pareto front) for each of the 5+10 critical objectives is shown in Figure \ref{fig:critical_objectives}. 

The whole database has $|\cP(\cS_H)| = 26$ bus-critical and 54 line-critical scenarios. While not all critical scenarios are recovered since Algorithm \ref{alg:dumbo} only gets exposed to less than half of the entire search space when it decides to stop, the overall number of critical scenarios is comparable and the number of function evaluations before stopping is stable across the runs. Furthermore, and most importantly, in each run, the magnitude of the identified max violations is almost identical, showing that while we do not identify exactly the same Pareto set, the overall Pareto front is nearly the same, and we find scenarios very close to the maximum violation in every run. This is a key take-away since, while we may not find the exact layout of PV adoption defined to be critical, the scenarios found lead to the same issues in grid equipment, making our methodology a reliable tool for upgrade planning. 
    
\begin{table}[]
    \centering
    \begin{tabular}{l|rrrr}
         & Simulated & Evaluated &  Critical & Critical  \\
        \textit{Run} & &  & Bus scens & Line scens \\
        \hline
        1 & 10600 & 392 &  16 & 33 \\ 
        2 & 14950 & 524 &  15 & 31 \\
        3 & 11500 & 398 &  11 & 29 \\ 
        4 & 13000 & 440 &  15 & 29 \\ \hline
    \end{tabular}
    \caption{Stability analysis of Algorithm \ref{alg:dumbo}. Each row is a different run with the search space constructed by independent sampling from a fixed database of $H=25000$ scenarios.}
    \label{tab:algo_stability}
\end{table}

\edit{Next, we perform a sensitivity analysis for four key parameters in Algorithm \ref{alg:dumbo}: the number of candidate scenarios $M$ at each BO step, the number of samples $N$ to compute $\alpha_{ND}$ in \eqref{eq:prob_crit}, the stopping threshold $\bar{\tau}$, and the number of new scenarios simulated at each BO step $N_{\text{expand}}$. To control for the intrinsic randomness due to scenario simulation, all the reported values are 
averages over 10 runs of Algorithm \ref{alg:dumbo}; we moreover list the respective standard deviations. Table \ref{tab:sensitivity_analysis} reports the total number of simulated scenarios, the number of evaluations (equivalent to computational time), and the number of found critical scenarios for the bus and line objectives. 
we report the average number of simulated, evaluated, bus-critical and line-critical scenarios. The discussion below is relative to the baseline parameters from Table \ref{tab:feeder}.} 

\edit{While the number of scenarios simulated by our algorithm is by design sensitive to above tuning parameters, robustness is reflected in the stable number of \emph{evaluated} scenarios, as well as a consistent number of found critical scenarios. Performance is little impacted by $N$, although setting it too low leads to a larger variance in all key metrics, i.e., a less stable performance, as the acquisition function becomes too noisy. The number of candidates in the stopping criterion, $M$, is also benign, though it must be sufficiently large to prevent the algorithm from terminating prematurely due to a lack of promising scenarios considered when computing the stopping criterion. The algorithm is most sensitive to $N_e=N_{\text{expand}}$ that controls the growth of the search space. Setting $N_{\text{expand}}$ too low causes the algorithm to exhaust the search space and stop prematurely.  Taking it too large, causes the algorithm to be overwhelmed by the number of potential candidates with the result that the stopping criterion decreases very slowly.  Finally, the threshold $\bar{\tau}$ directly controls the stopping rule, and a large $\tau$ causes early termination, missing some critical scenarios. }

\edit{In terms of violation magnitudes, we ran a statistical test that confirmed that the distribution of maximum found violations is statistically the same between the baseline and the variants in Table \ref{tab:sensitivity_analysis}, so the algorithm consistently finds the worst violations of each objective. Overall, the stable number of evaluations ($|\bX_{\Nstop}| \in [300,370]$) and of the critical scenarios ($14-16$ for buses $22-24$ for lines) discovered, indicates the robustness of our approach; the exception being the rows with $N=10$ (very large variance across runs) and $M=500$, $N_{\text{expand}}=50$ and $\bar{\tau}=0.5$ (insufficient number of evaluations leading to some critical scenarios missed).}


\begin{table}[!t]
\centering
    \resizebox{\columnwidth}{!}{
    $\begin{array}{lccrr}
    \toprule
     & \textrm{Simulated} & \textrm{Evaluated} & \textrm{Critical}  & \textrm{Critical} \\
     & |\cS_{\Nstop}| & |\bX_{\Nstop}| & \textrm{Bus scens} & \textrm{Line scens} \\
    \midrule
    N=10 & 9540 \pm 3590 & 362 \pm 138  & 14.3  \pm 4.0 & 23.3 \pm3.1 \\
    N=25 & 9000 \pm  219 & 338 \pm 69  & 15.2 \pm 3.8 & 22.9  \pm2.4 \\
    \mathbf{N=50} & 9480  \pm 1450 & 368 \pm 52 & 14.4 \pm 1.7 & 22.4  \pm 2.3 \\
    N=100 & 8740 \pm  148 & 330 \pm  54 & 14.0 \pm  1.9 & 21.7  \pm2.4 \\
    \midrule
    M=500 & 5763 \pm 898 & 216  \pm 37  & 14.8 \pm  2.7 & 22.4 \pm 4.2 \\
    M=1000 & 7900 \pm 1310 & 312 \pm 50 & 12.6 \pm  2.8 & 23.3 \pm 2.1 \\
    M=2000 & 8727 \pm 1150 & 342 \pm  48 & 15.0 \pm  3.4 & 23.4 \pm 2.8 \\
    \mathbf{M=3000} & 9480  \pm 1450 & 368 \pm 52 & 14.4 \pm 1.7 & 22.4  \pm 2.3 \\
    M=4000 & 9620 \pm 2200 & 361 \pm  72 & 14.4 \pm  3.5 & 23.2 \pm 2.8 \\
    \midrule
    N_e=50 & 3695 \pm 247 & 210 \pm  38 & 11.0 \pm  3.7 & 20.2 \pm 3.4 \\
    N_e=100 & 5845 \pm 540 & 322 \pm 39 & 12.6 \pm 3.0 & 22.4 \pm 3.6 \\
    \mathbf{N_e=200} & 9480  \pm 1450 & 368 \pm 52 & 14.4 \pm 1.7 & 22.4  \pm 2.3 \\
    N_e=300 & 12570 \pm 2360 & 369 \pm 68 & 17.2 \pm  3.9 & 23.6 \pm 2.7 \\
    N_e=400 & 16200 \pm 4590 & 378  \pm  93 & 17.7 \pm  3.4 & 24.6 \pm 4.0 \\
    \midrule
    \bar{\tau}=0.05 & 10360 \pm 2040 & 393 \pm  68 & 17.0 \pm  2.5 & 23.6 \pm 1.8 \\
    \mathbf{\bar{\tau}=0.1} & 9480  \pm 1450 & 368 \pm 52 & 14.4 \pm 1.7 & 22.4  \pm 2.3 \\
    \bar{\tau}=0.2 & 8491 \pm  950 & 328 \pm  38 & 15.2 \pm 2.1 & 23.6 \pm 3.2 \\
    \bar{\tau}=0.5 & 6818 \pm 1387 & 265 \pm  55 & 13.3 \pm 2.3 & 22.4 \pm 3.6 \\
    \bottomrule
    \end{array}$}
\caption{\edit{Sensitivity analysis of Algorithm \ref{alg:dumbo}. Each row corresponds to the average of 10 runs on a fixed database $\cS_H$ of $H=25000$ scenarios, with $\pm$ indicating the respective standard deviation across runs. Baseline parameters corresponding to the results in Table \ref{tab:feeder} are in bold. }} 
\label{tab:sensitivity_analysis}
\end{table}

\edit{Another algorithm parameter that ought to be mentioned is the number of bus groups $P$ that determines  the number of bus objectives. This parameter represents the number of voltage control areas in the network, which are often a small number and very specific of each feeder. Similarly, the violation threshold $\bar{s}$ (Line 5 of Algorithm \ref{alg:dumbo}) is another user-defined parameter that is network-dependent. One way to pick $\bar{s}$ is to look at the distribution of objective values on the set of initial evaluated scenarios.}

\section{Discussion}\label{sec:discuss}

\subsection{Surrogate Performance}

\begin{figure}[htb!]
    \centering
    \begin{tabular}{cc}
    $n=75$ & $n=333$ \\
    \includegraphics[width=0.24\textwidth]{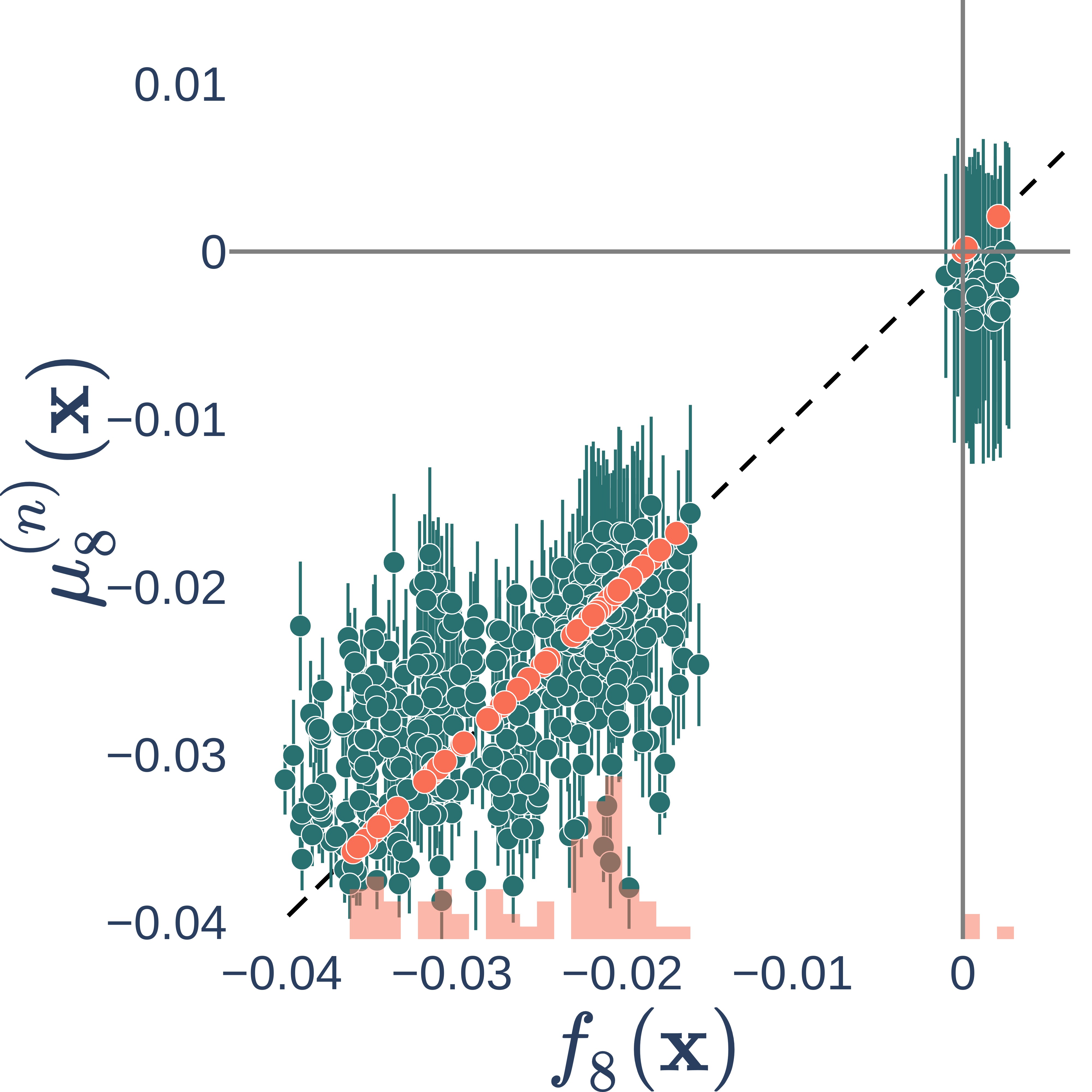}
    \hspace{-0.5cm} & 
    \includegraphics[width=0.24\textwidth]{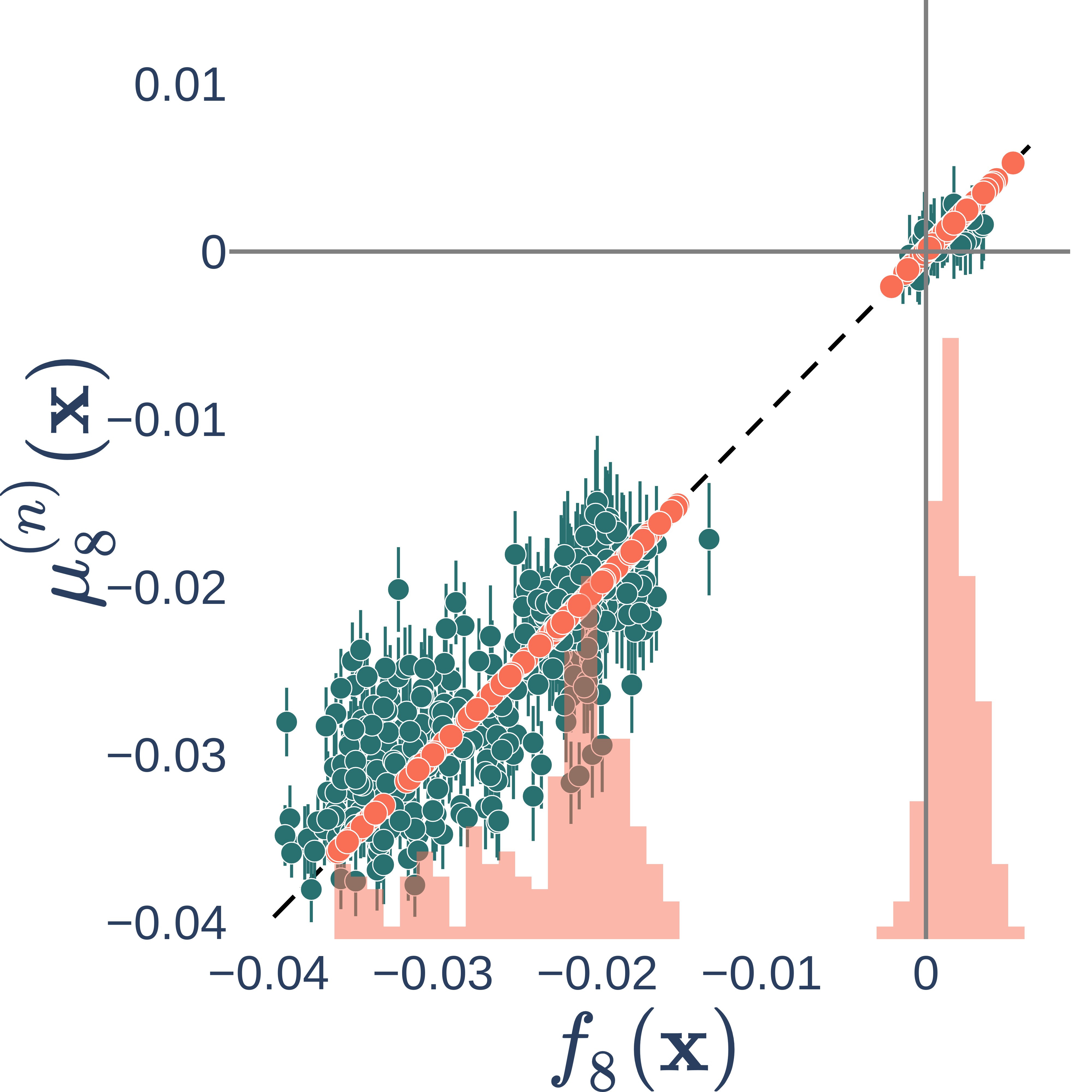} \\
    \end{tabular}
    \caption{True stress objective ($x$-axis) vs. GP prediction, i.e. posterior mean ($y$-axis) for the critical bus objective $f_8(\bx)$, after evaluating $n=75$ (\emph{left}) and $n=333$ scenarios (\emph{right}) respectively. Vertical error bars indicate $\pm 2$ standard deviations around the mean. Orange points are in-sample scenarios and green points are out-of-sample. The rug plots indicate the distribution of the training outputs $\{f(\bx), \bx \in \cD_n \}$.}
    \label{fig:gp_preds}
\end{figure}

It is instructive to illustrate
the evolution of the GP surrogates $\widehat{f}_k(\cdot)$ over the BO iterations $n=n_0,\ldots$. Because the search space $\cX$ is high-dimensional, we cannot directly visualize $\hat{f}^{(n)}_k(\cdot)$; instead we show in Figure \ref{fig:gp_preds} a scatterplot of the predicted mean (and standard deviations) $\mu^{(n)}_k(\cdot)$ objective values  against the true values $f_k(\cdot)$ at the initial step $n=n_0$ and at the last step $n=\Nstop$ when the algorithm terminated. 
The predictions are shown for the already evaluated scenarios $\cD_n$, as well as a (subset) of the current search set $\cS_n$. The uncertainty of the predictions is indicated by the vertical bars that span $\mu^{(n)}_k(\bx) \pm 2\sigma^{(n)}_k(\bx)$. Note that for $\bx \in \cD_n$, there is almost no uncertainty due to the small $\sigma_k^2$.

Accurate predictions would make $\mu^{(n)}_k(\bx)= f_k(\bx)$ lie on the diagonal.  As $n$ increases, two effects occur. First, the algorithm focuses on scenarios that are close to critical, i.e.~with $f_k(\bx) > 0$. Therefore, it shifts learning to the ``right'', cf.~the rug plots in the left and right panels. Second, for $n=333$ the GPs have more training data and therefore yield more accurate (smaller prediction errors) and more precise (smaller $\sigma^{(n)}_k(\bx)$) fits. In conjunction, these two effects shift the training observations and asymmetrically improve accuracy, \edit{as witnessed in} the right panel of Figure \ref{fig:gp_preds}. The much improved quality of $\hat{f}_k(\cdot)$  in the top-right quadrant at $n=333$ compared to initial fit at $n=75$ shows that  
the surrogates are adept at learning individual $f_k$'s in their critical regions as the BO progresses.

\subsection{Kernel Selection}
\edit{The performance of the GP surrogate in Algorithm \ref{alg:dumbo} hinges on the choice of its kernel. Standard kernels, such as the Matèrn or radial basis function (RBF),  assume continuous inputs and are ill-suited for binary vectors modeling our adoption scenarios. The Hamming distance we employ is the natural choice to measure distance between scenarios. An alternative is to employ embedding techniques to project the high-dimensional discrete $\bx$ into a lower-dimensional continuous space, where an RBF or Matèrn kernel can then be applied. Embedding can be done via random projection, for example the REMBO algorithm \cite{wang_bayesian_2016}; via dictionary embedding based on Hamming distance to an exogenously specified basis set, or via variational autoencoders.}


\edit{
Embedding is conceptually promising for our problem since
only a subset of adopters is expected to influence each bus or line objective. Thus, the effective dimensionality of our problem is likely much lower than the input dimension. Motivated by this, we compare our categorical kernel to two prominent embedding approaches. First, we implemented REMBO \cite{wang_bayesian_2016} and vary the embedding dimension and the resampling frequency of the random projection matrix, and test both RBF and Matèrn kernels. Second, we applied the dictionary embedding method (BODi) \cite{deshwal_bayesian_2023}, evaluating both the binary-wavelet dictionary and a  variant of our own, where at each step the scenarios in the Pareto set are used as basis vectors. For compactness, we report only the best-performing runs. Importantly, we emphasize that our use of these kernels is limited to surrogate modeling; we still evaluate our acquisition function on a subset of the search space.}

\begin{figure}
    \centering
    \includegraphics[width=\linewidth]{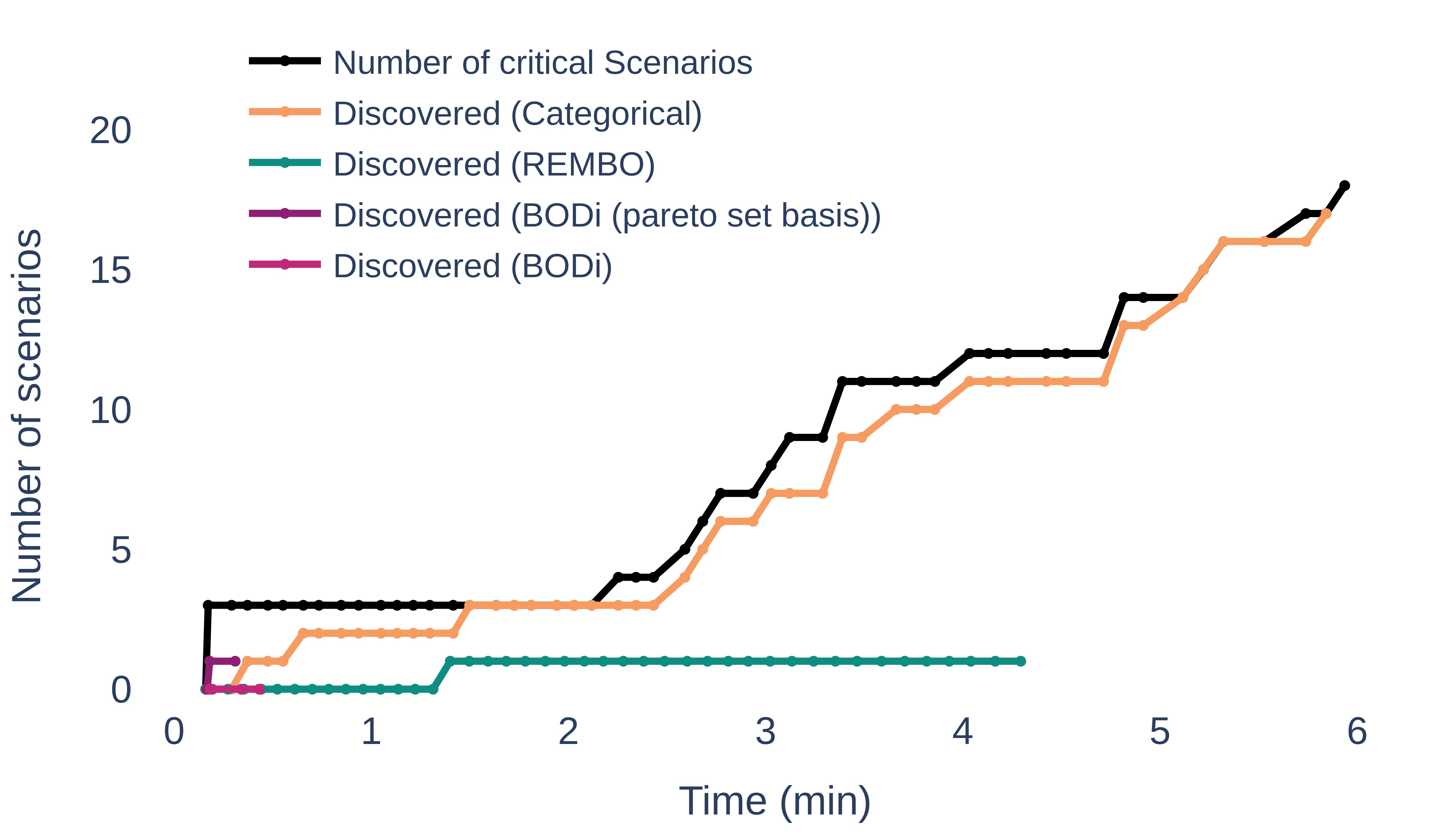}
    \caption{\edit{Number of bus-critical scenarios in the search space $\cS_n$ (black curve) and how many of those scenarios are identified by Algorithm \ref{alg:dumbo} using four different GP kernels, cf. Table \ref{tab:kernel_comp}. Each run is on the same search space but has its own stopping criterion, leading to different termination times. 
    }}
    \label{fig:kernel_comparison}
\end{figure}

\edit{Figure \ref{fig:kernel_comparison} reports the number of critical scenarios identified when running Algorithm \ref{alg:dumbo} with different kernels. REMBO was manually terminated after 500 evaluations  because its stopping criterion remained far from $\bar{\tau}$ and showed no indication of halting. BODi failed to recognize any scenarios as potentially critical and stopped prematurely after fewer than 50 evaluations, despite requiring a larger number of initial training points. For both REMBO and BODi, we fixed the embedding dimension at 20 (compared to $A=159$ for the categorical kernel), which substantially reduced GP fitting costs and which explains the shorter runtime for REMBO, despite having more function evaluations. 
The poor performance of REMBO and BODi stems from their apparent inability to capture the dependence of bus and line voltages on adoption scenarios. Table \ref{tab:kernel_comp} summarizes out-of-sample prediction errors across 500 scenarios, comparing our categorical kernel \eqref{eq:cat_kernel} the embedding-based alternatives. For every bus objective, the $\kappa_k$ in \eqref{eq:cat_kernel} achieves consistently lower RMSE and outperforms both REMBO and BODi. Similar results hold for line objectives, underscoring the advantage of directly modeling the binary structure of the input space. }

\begin{table}[ht]
    \centering
    \begin{tabular}{lrrrr}
    \toprule
     & REMBO & BODi (wavelet) & BODi (Pareto set) & Cat. (Eqn \eqref{eq:cat_kernel})\\
    \midrule
    $f_1$ & 1.62 & 1.12 & 1.17 & 0.43 \\
    $f_3$ & 1.63 & 0.99 & 0.95 & 0.41 \\
    $f_5$ & 0.94 & 0.6 & 0.58 & 0.36 \\
    $f_7$ & 0.8 & 0.69 & 0.65 & 0.15 \\
    $f_8$ & 1.57 & 1.01 & 1.12 & 0.33 \\
    \bottomrule
    \end{tabular}
    \caption{\edit{Predictive accuracy of the five critical bus objectives for different choices of GP kernels. We report predictive RMSE $(\times 10^{-2})$ over 500 unevaluated scenarios randomly selected from the search space after evaluating $n=333$ scenarios selected by Algorithm \ref{alg:dumbo}. 
    } }
\label{tab:kernel_comp}
\end{table}
\vspace{-1em}

\subsection{Most Relevant Adopters}\label{sec:most-relevant}
Our choice of the categorical GP kernel furnishes a tool to interpret adopter relevance, i.e.~to get information about which specific adopter buses are driving different bus and line stresses. Specifically, leveraging the separable nature of \eqref{eq:cat_kernel} the fitted relevance parameters $\pmb{\theta}_k$ indicate how much changes in the adoption pattern $\bx$ affect the GP predictions. High value for $\theta_{k,j}$ indicates that adoption by agent $j$ significantly impacts the stress in objective $k$. In that way, fitting the GP kernels as part of the BO provides the association between adopters and future grid stresses.

\begin{figure}[!htb]
    \centering
    \includegraphics[width=\linewidth]{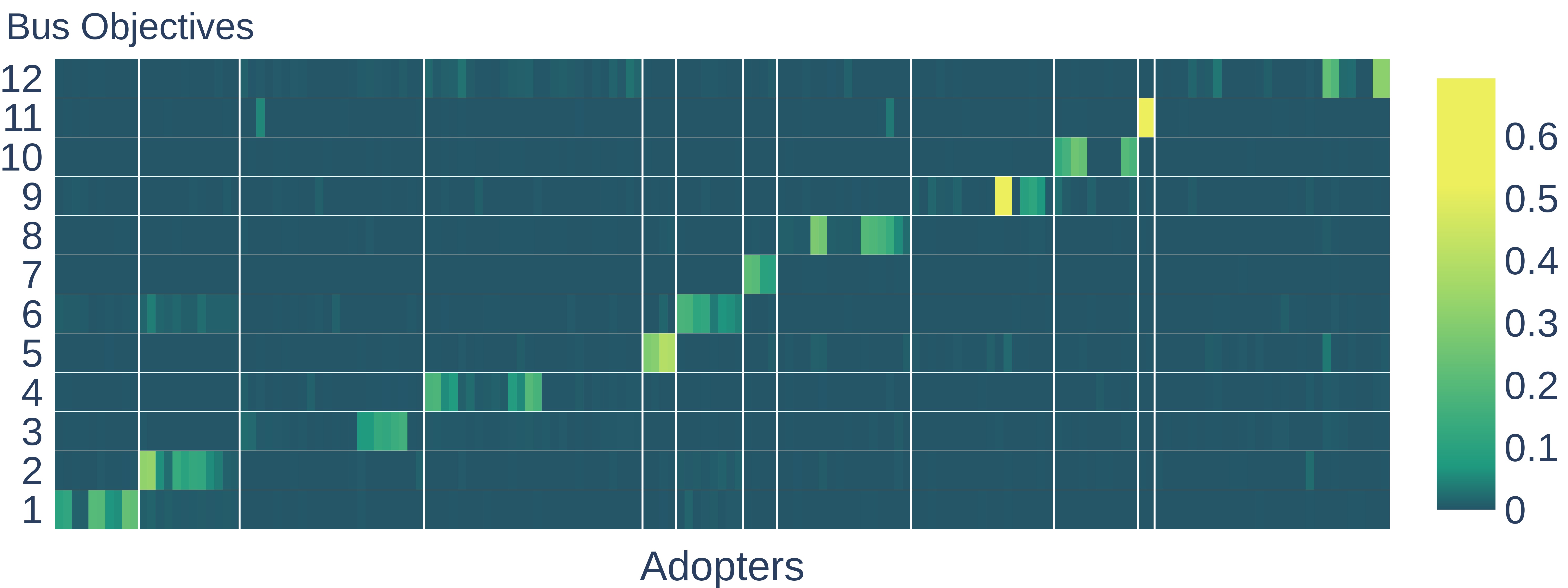}
    \caption{\edit{Heatmap of adopter relevance $\tilde{\theta}_{k, j}$ (brighter is higher) in feeder p4rhs8. Rows represent bus objectives $f_k$, $k=1, \ldots, {12}$, columns are the adopters $j=1,\ldots, 159$. Vertical white lines separate adopters according to the bus groups they belong to.}}
    \label{fig:ls_heatmap}
\end{figure}

We plot the relevance of each adopter $\tilde{\theta}_{k, j} := 1-e^{-\frac{\theta_{k, j}}{A}}$ in Figure \ref{fig:ls_heatmap} for feeder p4rhs8, with the bus objectives on the $y$-axis and the adopters on the $x$-axis, sorted by the partitions $\nodes_{p}$ to which they belong. The ARD feature of our GPs endogenously identifies the relevant adopters for each objective, shown by the brighter colors. The sparsity of the heatmap implies that 
%
for each bus objective, only a few adopters are relevant and there is a low-dimensional subspace of $\cX$ with ``active" dimensions. This sparsity helps the GPs to focus on these dimensions to model the stress values.
The right panel of Figure \ref{fig:p3uhs27} visualizes $\tilde{\theta}_{1, j}$ for objective $f_1$ (based on buses indicated by the orange points) for feeder p3uhs27 spatially overlaid on the grid network. 
%
We find that 
the resulting most-relevant ``active'' adopters are consistent with grid topology, corresponding to buses that are in the cluster $\mathcal{B}_1$ defining $f_1$, or to nearby adopters with large PV capacities. 

\edit{The identification of these relevant adopters provides utilities with a group of consumers that are likely to trigger grid investments through their adoption decisions. This information can be used to schedule grid upgrades in response to PV interconnection requests from this group. It can also offer useful insights on where to incentivize load growth or procure local flexibility services to defer such upgrades.}

\section{Conclusions and Future Work}
In this work, we propose a new methodology to efficiently search through thousands of scenarios of PV adoption to identify those that lead to worst violations of bus voltages and line flows. Our Bayesian Optimization-driven search accurately navigates the high-dimensional scenario space, dramatically improving ($27-52\times$ fewer evaluations) over a brute force approach that evaluates all simulated scenarios. We also demonstrate that the search is highly nontrivial and that naive heuristics like sorting by aggregate adopted PV capacity will miss most critical scenarios.

\edit{Several aspects of our methodology should be investigated further. Further analysis of the best GP surrogates could offer faster initialization and re-fitting of the $f_k$'s; for example one could try to leverage the latent lower-dimensional structure hinted in Section \ref{sec:most-relevant}. It might be worthwhile to incorporate correlation among GP surrogates to fuse learning of related stress objectives. Another idea is to fit GPs directly for individual bus voltages $volt_b(\bx)$ in \eqref{eq:stress_bus}, rather than on groups of buses $\nodes_k$, and then group predictions. Some preliminary experiments showed promising gains in accuracy, although at a higher computational cost.  Alternative acquisition functions might be warranted for very large feeders (over 1000 nodes) where $\alpha_{ND}$ becomes slow.}

\edit{The present work focused on distribution planning for radial grids. Further analysis is warranted for larger and/or meshed grids where more lines are likely to be critical. Finally, one could extend the methodology to consider multi-horizon planning where DER adoption evolves over multiple time steps and criticality is defined both locationally (what is violated) and temporally (when).}


\bibliographystyle{IEEEtran}
\bibliography{References}

\end{document}